%% file: mtsummit25.tex
\pdfoutput=1

\documentclass[11pt]{article}

\usepackage[final]{mtsummit25}

\usepackage{times}
\usepackage{latexsym}

\usepackage{booktabs}
\usepackage{graphicx}
\usepackage{threeparttable}
\usepackage{array}
\usepackage{adjustbox}
\usepackage{boldline}
\usepackage{makecell}
\usepackage{graphicx, subfig}
\usepackage{float}
\usepackage{xcolor}
\usepackage{mdframed}
\usepackage{tcolorbox}
\usepackage{tikz}
\usepackage{comment}
\usepackage{enumitem}
\usepackage{float}
\usepackage{booktabs}   
\usepackage{adjustbox}
\usepackage{threeparttable} 
\usepackage{caption}
\usepackage{hyperref}

\usepackage{copyright}

\usepackage[T1]{fontenc}

\usepackage[utf8]{inputenc}

\usepackage{microtype}

\usepackage{inconsolata}

\usepackage{graphicx}
\usepackage{xcolor}
\usepackage{multirow}
\usepackage{array}
\newcommand{\ev}[1]{{\textcolor{teal}{#1}}}

\title{Are We Paying Attention to \textit{Her}?\\ Investigating Gender Disambiguation and Attention \\in Machine Translation} 

\author{
Chiara Manna\hspace{2em}Afra Alishahi\hspace{2em}Frédéric Blain\hspace{2em}Eva Vanmassenhove \\
[0.5em]
\texttt{\{c.manna, a.alishahi, f.l.g.blain, e.o.j.vanmassenhove\}@tilburguniversity.edu} \\
[0.5em]
CSAI, Tilburg University \\
[0.5em]
Netherlands
}

\begin{document}
\maketitle
\begin{abstract}
While gender bias in modern Neural Machine Translation (NMT) systems has received much attention, the traditional evaluation metrics for these systems do not fully capture the extent to which models integrate contextual gender cues. We propose a novel evaluation metric called Minimal Pair Accuracy (MPA) which measures the reliance of models on gender cues for gender disambiguation. We evaluate a number of NMT models using this metric, we show that they ignore available gender cues in most cases in favour of (statistical) stereotypical gender interpretation. We further show that in anti-stereotypical cases, these models tend to more consistently take male gender cues into account while ignoring the female cues. Finally, we analyze the attention head weights in the encoder component of these models and show that while all models to some extent encode gender information, the male gender cues elicit a more diffused response compared to the more concentrated and specialized responses to female gender cues.\footnote{The code used in this work is made publicly available at \href{https://github.com/chiaramanna/gender-cue-integration-MT}{github.com/chiaramanna/gender-cue-integration-MT}.}
\end{abstract}
\section{Introduction}
\input{notes/intro_EVA}

Our evaluation reveals that the assessed NMT models do not consistently leverage the contextual gender cues provided. Instead, they often seem to revert back to statistical (and thus stereotypical) patterns rather than context. We furthermore observe a discrepancy between the integration of masculine versus feminine cues. The presence of a masculine pronoun with pro-stereotypically female professions often enables the model to correctly infer the gender of a lexically gender-ambiguous target word while the reverse does not hold. Additionally, our analysis of attention head weights in the encoder component indicates that, although all models encode gender information to some extent, masculine cues elicit a more diffused response, whereas feminine ones generate more concentrated and specialized attention patterns.


\section{Bias Statement}

We define gender bias in MT as the tendency of models to default to learned statistical associations rather than systematically relying on contextual information for gender disambiguation. We focus on cases where gender is unambiguously expressed in the source sentence -- typically through pronouns referring to human entities -- capturing one subtype of gender bias. Ambiguous cases -- lacking explicit gender cues -- fall outside the scope of this paper.

While our framework targets the English-Italian (EN--IT) language pair, it is broadly applicable to any setting where gender must be explicitly marked in the target language. We particularly highlight stereotypical bias, for which models successfully generate feminine translations when the target word (\textit{i.e.}, the profession noun) is already associated with women (\textit{e.g.} \textit{librarian} $\rightarrow$ \textit{bibliotecaria}), but struggle to override male defaults in anti-stereotypical contexts. This asymmetry suggests that gender disambiguation might be driven by learned priors rather than syntactic dependencies, reinforcing a male-as-norm bias \cite{danesi-2014}. Such bias can lead to both representational harm, by perpetuating traditional gender roles, and allocational harm, by systematically underrepresenting women in male-dominated professions \citep{blodgett-etal-2020-language}.

Our analysis only considers binary gender due to the constraints of the WinoMT dataset, which relies on U.S. Labor Statistics and morphological analysis tools that categorize gender along a binary axis. While we acknowledge that this is a major limitation and gender is not a binary construct, there is no standardized approach to systematically evaluate non-binary gender bias in MT. Broader inclusivity challenges persist and underscore the need for future work to develop more inclusive methodologies that better reflect gender as a spectrum.

\section{Related Work}\label{sec:rw}
\input{notes/related_work_EVA}

\section{Experimental Setup}

In order to examine the extent to which contextual gender cues contribute to the representation of profession nouns for different models, we analyzed how multiple state-of-the-art models (Section~\ref{subsec:models}) integrate contextual gender cues provided in the WinoMT challenge set in the gender disambiguation process (Section~\ref{subsec:data}). 


\subsection{Models}\label{subsec:models}
We investigate three pre-trained encoder-decoder models for English-to-Italian translation, selecting them based on their widespread use and high ranking among open source translation models on Hugging Face\footnote{\url{https://huggingface.co/}}, allowing for greater transparency in analyzing their internal mechanisms. While we focus on encoder-decoder models, the framework can be extended to encoder-only or decoder-only architectures, adopted by LLMs.\\[\medskipamount]
\textbf{OPUS-MT EN--IT}\footnote{\href{https://huggingface.co/Helsinki-NLP/opus-mt-en-it}{huggingface.co/Helsinki-NLP/opus-mt-en-it}} \citep{opus_mt} is a bilingual model specifically trained for English-to-Italian translation using supervised learning on parallel corpora from the OPUS dataset \citep{opus_data}. It consists of \texttt{6} encoder layers,  \texttt{6} decoder layers, and  \texttt{8} attention heads per layer. \\[\medskipamount]
\textbf{NLLB-200}\footnote{\href{https://huggingface.co/facebook/nllb-200-distilled-600M}{huggingface.co/facebook/nllb-200-distilled-600M}} \citep{nllb200} is a multilingual model trained to support 200 languages. We make use of the distilled version, which contains \texttt{12} encoder layers, \texttt{12} decoder layers, and \texttt{16} attention heads per layer, with \texttt{600M} parameters. The model is trained with a combination of supervised and self-supervised learning on multilingual corpora. \\[\medskipamount]
\textbf{mBART}\footnote{\href{https://huggingface.co/facebook/mbart-large-50-many-to-many-mmt}{huggingface.co/facebook/mbart-large-50-many-to-many-mmt}} \citep{liu-etal-2020-multilingual-denoising} is a denoising \texttt{autoencoder}-based multilingual model designed for cross-lingual tasks, including translation. Unlike the previous models, which rely primarily on parallel corpora, mBART is first pretrained on monolingual text using a denoising autoencoding objective, where the model learns to reconstruct text from a corrupted version. As NLLB-200, it features \texttt{12} encoder layers, \texttt{12} decoder layers, and \texttt{16} attention heads per layer.

\subsection{Data}\label{subsec:data}
We use the \textbf{WinoMT} challenge set \citep{winomt}, which was designed to evaluate gender bias in MT systems from English to 8 target languages with grammatical gender. It includes a regular set of 3888 synthetic sentences derived from Winogender \citep{rudinger-etal-2018-gender} and WinoBias \citep{zhao-etal-2018-gender} -- two benchmark datasets for coreference resolution -- in which a primary entity (\textit{e.g.}, a profession noun) and a pronoun are in a coreference relation. Instances are balanced for masculine and feminine genders, as well as pro- and anti-stereotypical gender-role assignment based on U.S. Labor Statistics \citep{zhao-etal-2018-gender}. 

\begin{figure}[h!]
    \centering

    \begin{tcolorbox}[
        colback=red!4, colframe=red!50, 
        boxrule=1pt, arc=6pt,
        left=4pt, right=4pt, top=4pt, bottom=4pt
    ]
        \textbf{PRO-S:} \hspace{0.5em} The chief gave the \textbf{housekeeper} a tip because \textcolor{red}{\textbf{she}} was helpful.
        
    \end{tcolorbox}
    \begin{tcolorbox}[
        colback=blue!4, colframe=blue!50, 
        boxrule=0.8pt, arc=8pt, 
        left=6pt, right=6pt, top=4pt, bottom=4pt 
    ]
        \textbf{ANTI-S:} \hspace{0.5em} The chief gave the \textbf{housekeeper} a tip because \textcolor{blue}{\textbf{he}} was helpful.
    \end{tcolorbox}
    \caption{\small Example of a pro-stereotypical (PRO-S) and anti-stereotypical (ANTI-S) gender role assignment from the WinoMT challenge set.}\label{fig_2:pro_anti}
\end{figure}

Additionally, two sets of 1584 instances each are provided -- \texttt{en\_pro} and \texttt{en\_anti} -- where the same profession nouns are paired with pronouns based on pro- and anti-stereotypical gender-roles, respectively. To illustrate this, we present the same sentence from both sets in Figure~\ref{fig_2:pro_anti}. In the pro-stereotypical sentence (PRO-S), the gender of \textit{housekeeper} aligns with the gender that most often carries out this particular profession according to the U.S. Bureau of Labour Statistics.\footnote{In 2024, 87.7\% of housekeepers are women -- see: \href{https://www.bls.gov/cps/cpsaat11.htm}{bls.gov/cps/cpsaat11.htm}.} Conversely, in the anti-stereotypical (ANTI-S) setting the gender role assigned can be considered more challenging as statistically\footnote{Again, based on US statistics.} men are less likely to carry out the job of \textit{housekeeper}.

\begin{table}[t!]
\centering
    \small 
    \begin{threeparttable}
        \begin{adjustbox}{max width=\linewidth}
        \begin{tabular}{llccc}
            \toprule
            \textbf{Set} & \textbf{Model} & \textbf{Overall} & \textbf{Male} & \textbf{Female} \\
            \midrule
            \multirow{3}{*}{REG} 
                & OPUS\_MT & 42.6\% & 70.1\% & 20.6\% \\
                & NLLB-200 & 57.0\% & 79.6\% & 41.8\% \\
                & mBART    & \textbf{60.9\%} & \textbf{83.2\%} & \textbf{46.5\%} \\
            \midrule
            \multirow{3}{*}{PRO-S} 
                & OPUS\_MT & 55.7\% & 77.3\% & 34.1\% \\
                & NLLB-200 & 74.9\% & 87.4\% & \textbf{62.5\%} \\
                & mBART    & \textbf{76.6\%} & \textbf{92.2\%} & 61.0\% \\
            \midrule
            \multirow{3}{*}{ANTI-S} 
                & OPUS\_MT & 34.2\% & 59.1\% & 9.2\% \\
                & NLLB-200 & 47.3\% & 70.4\% & 24.2\% \\
                & mBART    & \textbf{54.0\%} & \textbf{71.9\%} & \textbf{35.9\%} \\
            \bottomrule
        \end{tabular}
        \end{adjustbox}
        \vspace{0.5mm}
        \caption{\small Overall, male and female accuracy on WinoMT for the OPUS\_MT, NLLB-200, and mBART models on the regular (REG), pro-stereotypical (PRO-S), and anti-stereotypical (ANTI-S) sets.}
        \label{tab:gender_accuracy}
    \end{threeparttable}
\end{table}


\section{Evaluating Context Integration in Gender Disambiguation}\label{sec:Eval_MPA}

In this section, we will first delve into the evaluation of contextual cue integration through our novel metric. Next, in Section~\ref{sec:contextattention}, we continue with the analysis of the encoder attention head weights to investigate how gender cues are integrated into the target representations.

\subsection{WinoMT Evaluation}
WinoMT provides an integrated evaluation pipeline that relies on automatic word alignment and morphological analysis to extract the grammatical gender of the primary entity from each translated sentence. Comparing the extracted gender information with the gold label enables us to compute three accuracy measures: 
\begin{description}[noitemsep]
    \item[Overall Accuracy:] Percentage of correctly gendered entities.
    \item[Male Accuracy:] Accuracy for entities with a masculine gold label.
    \item[Female Accuracy:] Accuracy for entities with a feminine gold label.
\end{description}

Table~\ref{tab:gender_accuracy} presents the gender accuracy for all models across the regular, pro- and anti-stereotypical sets. First of all, we observe that all models consistently perform better for: (i)  masculine referents and (ii) in stereotypical settings where the gender aligns with the societal expectations. When comparing the models, mBART outperforms NLLB-200 and OPUS-MT on all three sets (regular, stereotypical and anti-stereotypical) in terms of overall and male accuracy. Only on the stereotypical set, NLLB-200 (62.5\%) slightly outperforms BART (61.0\%) in terms of accuracy for female referents. 


\subsection{Minimal Pair Accuracy}\label{subsec:mpa}
While the aforementioned standard metrics provide an overall performance measure, they do not assess whether models effectively leverage contextual gender cues to resolve gender ambiguity during translation. In an attempt to move beyond these rather surface-level accuracy scores, we introduce and analyze the \textbf{Minimal Pair Accuracy (MPA)}, a novel metric designed to assess the degree to which models are able to effectively leverage contextual gender cues for disambiguation during translation. 

\begin{table}[t!]
    \centering
    \begin{tabular}{lc}
        \toprule
        \textbf{Model} & \textbf{MPA} \\ 
        \midrule
        OPUS\_MT  & 6.12\% \\
        NLLB-200 & 30.24\% \\
        mBART    & \textbf{38.45\%} \\
        \bottomrule
    \end{tabular}
    \vspace{0.5mm}
    \caption{\small MPA presents the percentage of cases where the model correctly disambiguated the source word based on the contextual gender cue.} 
    \label{tab:minimal_pair_accuracy}
\end{table}

We construct minimal pairs by extracting \texttt{PRO-S} and \texttt{ANTI-S} sentences, with the only distinguishing factor being the gendered pronoun. This allows us to analyze the effect of the gender cue on translation. For instance, as shown in Figure~\ref{fig:minimal_pair_example}, the profession noun \textit{librarian} is translated into Italian as either \textit{la bibliotecaria} (feminine) or \textit{il bibliotecario} (masculine), depending on the pronoun (e.g., \textit{she} or \textit{he}) in the English source sentence. We then focus on pairs where the model successfully disambiguates the profession noun in both cases by utilizing the contextual gender information provided by the pronoun. This way, \textbf{MPA} can give us an indication of how consistently the models integrate contextual gender cues.

\begin{figure*}[h!]
    \centering
    \begin{minipage}{0.48\textwidth}
       
        \begin{tcolorbox}[
            colback=red!4, colframe=red!50, 
            boxrule=1pt, arc=6pt,
            left=4pt, right=4pt, top=4pt, bottom=4pt
        ]
            \textbf{PRO-S:}\hspace{0.5em} The analyst consulted with the \textbf{librarian} because \textcolor{red}{\textbf{she}} knows a lot about books.
        \end{tcolorbox}

        \vspace{2mm} 

        \begin{tcolorbox}[
            colback=green!4, colframe=green!50, 
            boxrule=1pt, arc=6pt,
            left=4pt, right=4pt, top=4pt, bottom=4pt
        ]
            \textbf{OPUS-MT:}\hspace{0.5em} L'analista si è consultato con \textcolor{red}{\textbf{la bibliotecaria}} perché sa molto sui libri.
        \end{tcolorbox}

        \begin{tcolorbox}[
            colback=green!4, colframe=green!50, 
            boxrule=1pt, arc=6pt,
            left=4pt, right=4pt, top=4pt, bottom=4pt
        ]
            \textbf{NLLB-200:}\hspace{0.5em} L'analista ha consultato \textcolor{red}{\textbf{la bibliotecaria}} perché sa molto di libri.
        \end{tcolorbox}

        \begin{tcolorbox}[
            colback=green!4, colframe=green!50, 
            boxrule=1pt, arc=6pt,
            left=4pt, right=4pt, top=4pt, bottom=4pt
        ]
            \textbf{mBART:}\hspace{0.5em} L'analista ha consultato \textcolor{red}{\textbf{la bibliotecaria}} perché sa molto sui libri. 
        \end{tcolorbox} 
    \end{minipage}
    \hfill
    \begin{minipage}{0.48\textwidth}

        \begin{tcolorbox}[
            colback=blue!4, colframe=blue!50, 
            boxrule=1pt, arc=6pt,
            left=4pt, right=4pt, top=4pt, bottom=4pt
        ]
            \textbf{ANTI-S:}\hspace{0.5em} The analyst consulted with the \textbf{librarian} because \textcolor{blue}{\textbf{he}} knows a lot about books.
        \end{tcolorbox}

        \vspace{2mm}

        \begin{tcolorbox}[
            colback=green!4, colframe=green!50, 
            boxrule=1pt, arc=6pt,
            left=4pt, right=4pt, top=4pt, bottom=4pt
        ]
            \textbf{OPUS-MT:}\hspace{0.5em} L'analista si è consultato con  \textcolor{blue}{\textbf{il bibliotecario}} perché sa molto sui libri.
        \end{tcolorbox}

        \begin{tcolorbox}[
            colback=green!4, colframe=green!50, 
            boxrule=1pt, arc=6pt,
            left=4pt, right=4pt, top=4pt, bottom=4pt
        ]
            \textbf{NLLB-200:}\hspace{0.5em} L'analista ha consultato \textcolor{blue}{\textbf{il bibliotecario}} perché sa molto di libri. 
        \end{tcolorbox}

        \begin{tcolorbox}[
            colback=green!4, colframe=green!50, 
            boxrule=1pt, arc=6pt,
            left=4pt, right=4pt, top=4pt, bottom=4pt
        ]
            \textbf{mBART:}\hspace{0.5em} Il analista ha consultato \textcolor{blue}{\textbf{il bibliotecario}} perché sa molto sui libri.
        \end{tcolorbox}
    \end{minipage}
    \vspace{2mm}

    \caption{\small Example of accurate minimal pair translations constructed from the WinoMT challenge set. The left side (pro-stereotypical) assigns the feminine pronoun \textit{she} to the profession \textit{librarian}, while the right side (anti-stereotypical) replaces it with the masculine pronoun \textit{he}. The Italian translations correctly adapt the grammatical gender (\textit{la bibliotecaria} \textit{vs.} \textit{il bibliotecario}) across all models. Therefore, this pair contributes positively to the Minimal Pairs Accuracy (MPA) for each model.}
    
    \label{fig:minimal_pair_example}
    
\end{figure*}

The overall low MPA results presented in Table~\ref{tab:minimal_pair_accuracy} indicate that models struggle to consistently leverage contextual gender cues for disambiguation. However, NLLB-200 and mBART perform notably better, with an accuracy of 30.24\% and 38.45\% -- respectively -- as compared to OPUS-MT's significantly lower 6.12\%. 
\begin{table}[t!]
    \centering
    \begin{tabular}{lcc}
        \toprule
        \textbf{Model} & \textbf{Pro-F} & \textbf{Pro-M} \\
        \midrule
        OPUS\_MT  & \textbf{82.29\%} & 17.71\% \\
        NLLB-200 & 69.10\% & 30.90\% \\
        mBART    & 61.90\% & \textbf{38.10\% }\\
        \bottomrule
    \end{tabular}
    \vspace{0.5mm}
    \caption{\small A breakdown of the MPA. \textbf{Pro-F} refers to the percentage of correctly disambiguated minimal pairs where the profession would stereotypically be associated with women. \textbf{Pro-M} refers to the ones where the profession would stereotypically be associated with men.}
    \label{tab:gender_comp_mpa}
\end{table}

A closer examination of those accurate minimal pairs reveals yet another layer of asymmetry. 
Table~\ref{tab:gender_comp_mpa} presents the percentage of accurately translated minimal pairs where the profession is stereotypically associated with women (Pro-F) versus those where the profession is stereotypically associated with men (Pro-M). The results show that correctly disambiguating a profession noun based on a gender cue is much easier when the profession is stereotypically associated with women. In other words, stereotypical female professions are relatively easy to override with a masculine cue.

An example can be found in Figure 3, where all models correctly disambiguate a stereotypically female profession \textit{librarian}\footnote{In 2024, based on the US Labor Force Statistics, 89.2\% of librarians are women -- see: \href{https://www.bls.gov/cps/cpsaat11.htm}{bls.gov/cps/cpsaat11.htm}.} in both a stereotypical (PRO-S) and anti-stereotypical (ANTI-S) setting. Even in the ANTI-S condition, where the context provides a masculine cue (\textit{he}), the correct anti-stereotypical masculine form \textit{il bibliotecario} is generated by all three models. Overriding a stereotypically male-dominated profession is more difficult for all three models. When \textit{mechanic} -- a profession predominantly held by men\footnote{In 2024, based on the US Labor Force Statistics, only 3.2\% of mechanics are women -- see: \href{https://www.bls.gov/cps/cpsaat11.htm}{bls.gov/cps/cpsaat11.htm}.} -- is paired with \textit{she}, none of the models succeed in generating the expected feminine form, \textit{la meccanica}.



Specifically, OPUS\_MT shows that only 17.71\% of accurate minimal pairs successfully utilize an anti-stereotypical context to disambiguate a feminine referent. While this percentage increases slightly with the other models, it remains below 40\%, indicating a general difficulty in overriding male defaults.

These findings indicate that feminine cues only trigger gender disambiguation when the profession noun they refer to is stereotypically associated with the feminine gender. Otherwise, the investigated models often default to masculine terms, reinforcing an inherent male-as-norm bias \cite{danesi-2014}. Previous work supports this pattern, showing that language models, in fact, tend to follow a default-to-masculine reasoning process when assigning gender \citep{jumelet-etal-2019-analysing}. 


\section{Investigating Context Integration Through Attention}\label{sec:contextattention}
To gain further insight into how contextual gender information is encoded within Transformer models, we further investigate the extent to which gender cues are integrated into the representation of target words. 
For example, if a model correctly translates both {\bf PRO-S} and {\bf ANTI-S} examples in Figure~\ref{fig:minimal_pair_example}, we expect the representation of the target word {\it librarian} to be heavily influenced by the gender cue {\it she/he} in the original sentence. 
More specifically, we are interested in analyzing whether the attention mechanism contributing to the input representation of the target word attends to the gender cue and, if so, whether there are specific attention layers and heads that specialize in encoding gender cues.

\subsection{Setup}
Contextual information is leveraged through a multi-head attention mechanism in Transformer models. This operates at three levels in encoder-decoder architectures: self-attention in the encoder, self-attention in the decoder and cross-attention between the decoder and encoder representations~\citep{attention2017}. Previous work on context mixing in Transformer models has shown that encoder-only models effectively integrate contextual cues in their representations, while encoder-decoder models seem to relegate this task to the decoder \citep{, mohebbi-etal-2023-homophone, mohebbi-etal-2023-quantifying}. However -- in our setup -- the gender cue (\textit{i.e.}, the pronoun) is preceded by the target word (\textit{i.e.}, the profession noun) (see Figure \ref{fig_1:gender_bias_example}). As a result, decoder self-attention cannot account for it, as it only captures dependencies within already-generated tokens. 
Therefore, we focus on the self-attention patterns observed within the encoder in our analysis.\footnote{The analysis of cross-attention heads did not reveal notable patterns, but 
for the sake of completeness, the full set of cross-attention results are reported in the Appendix.}

\begin{figure}[htbp]
    \centering

    \begin{minipage}{0.45\textwidth}
        \centering
        \subfloat[OPUS-MT]{\includegraphics[width=\linewidth]{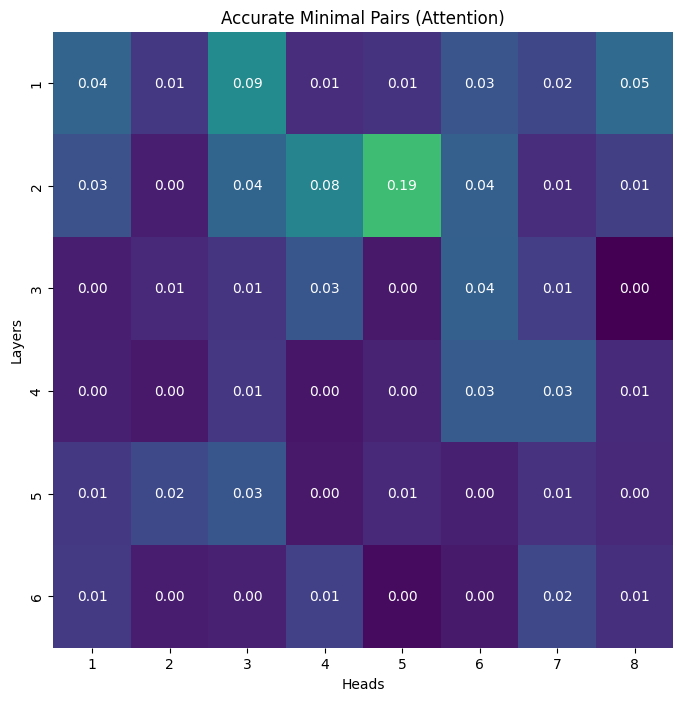}}
    \end{minipage}

    \vspace{2mm} 

    \begin{minipage}{0.45\textwidth}
        \centering
        \subfloat[NLLB-200]{\includegraphics[width=\linewidth]{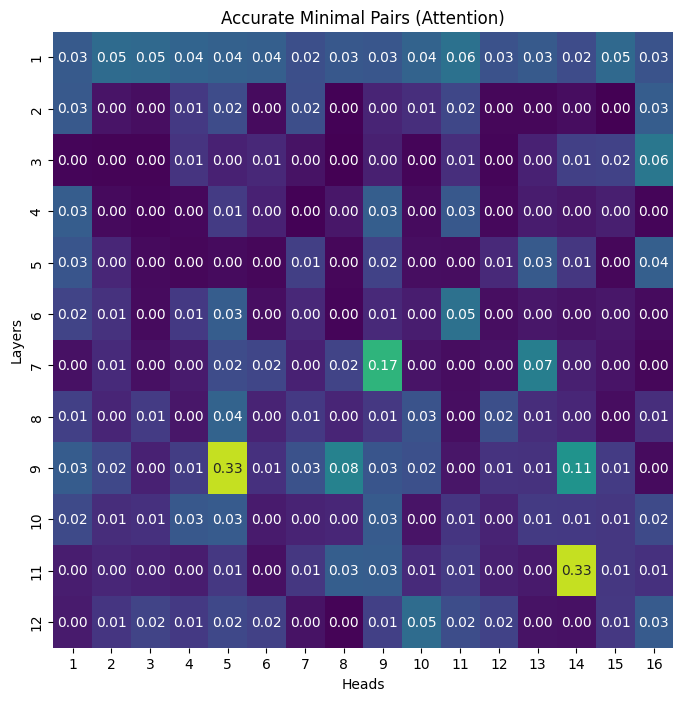}}
    \end{minipage}

    \vspace{2mm} 

    \begin{minipage}{0.45\textwidth}
        \centering
        \subfloat[mBART]{\includegraphics[width=\linewidth]{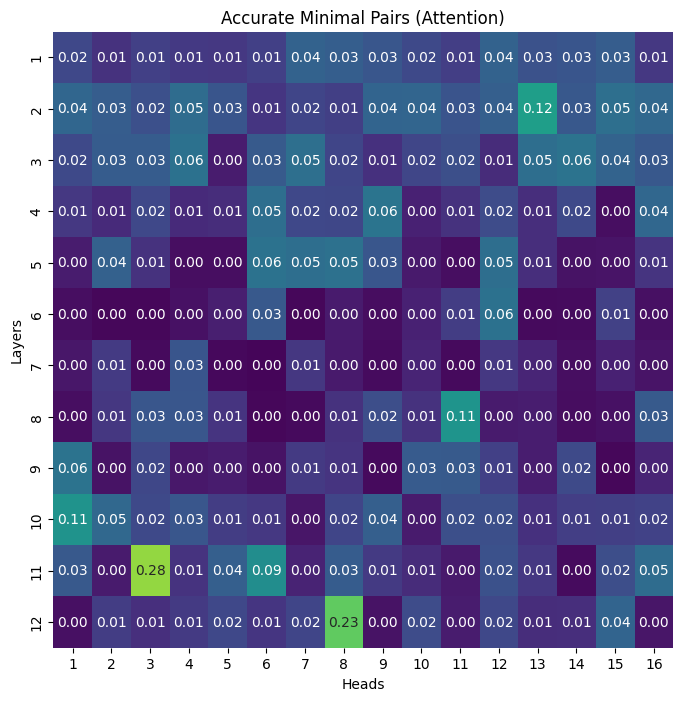}}
    \end{minipage}
    
    \vspace{2mm}
     
    \caption{\small{Heatmaps illustrating average encoder self-attention weights between the gender cue (\textit{i.e.}, pronoun) and the profession noun across accurate minimal pairs for each model. A standardized colormap is applied across all heatmaps.}}
    \label{fig:context_mix_1}
\end{figure}

\begin{figure*}[htbp]
    \centering
    \begin{minipage}{0.45\textwidth}
        \centering
        \subfloat[OPUS-MT]{\includegraphics[width=\linewidth]{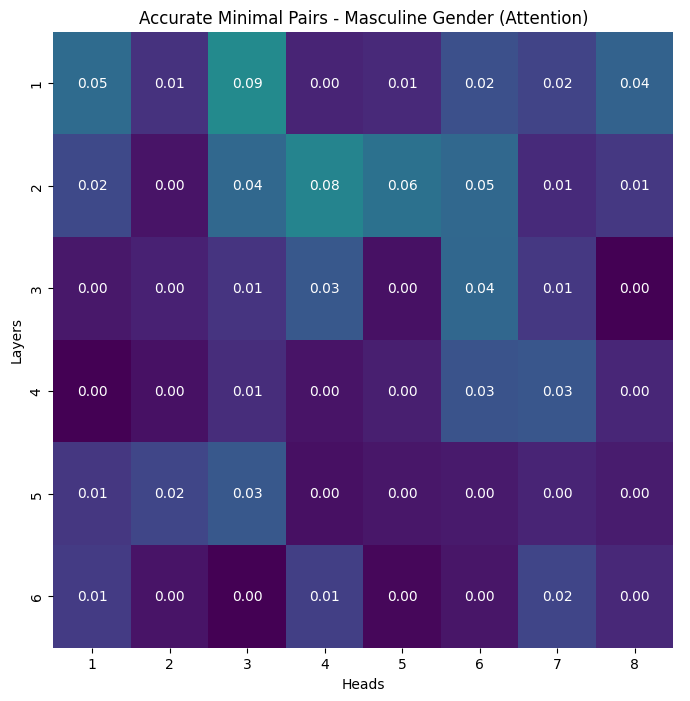}}
    \end{minipage}
    \hspace{-2mm}
    \begin{minipage}{0.45\textwidth}
        \centering
        \subfloat[OPUS-MT]{\includegraphics[width=\linewidth]{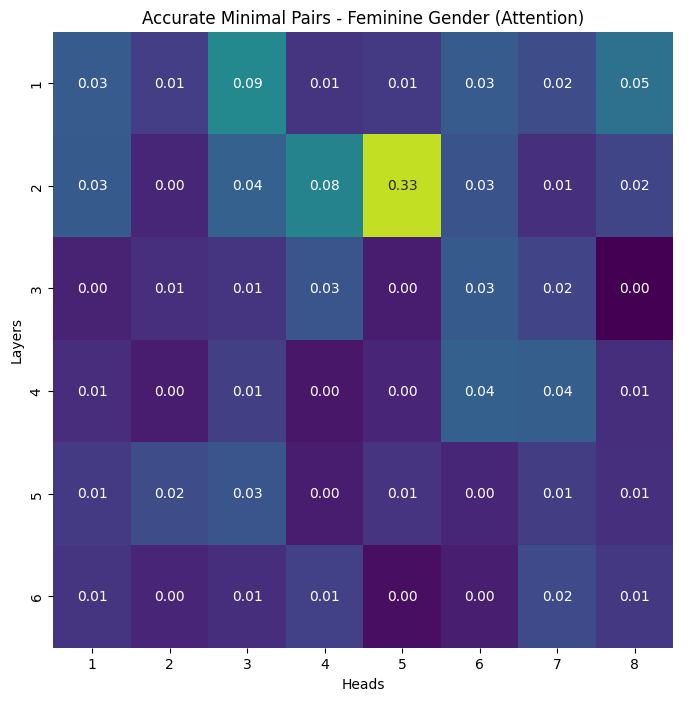}}
    \end{minipage}

    \vspace{2mm} 
    
    \begin{minipage}{0.45\textwidth}
        \centering
        \subfloat[NLLB-200]{\includegraphics[width=\linewidth]{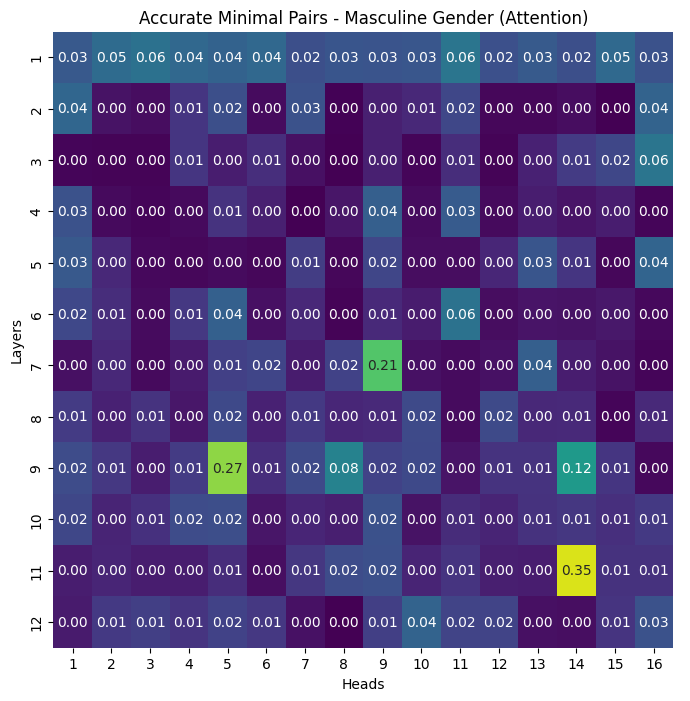}}
    \end{minipage}
    \hspace{-2mm}
    \begin{minipage}{0.45\textwidth}
        \centering
        \subfloat[NLLB-200]{\includegraphics[width=\linewidth]{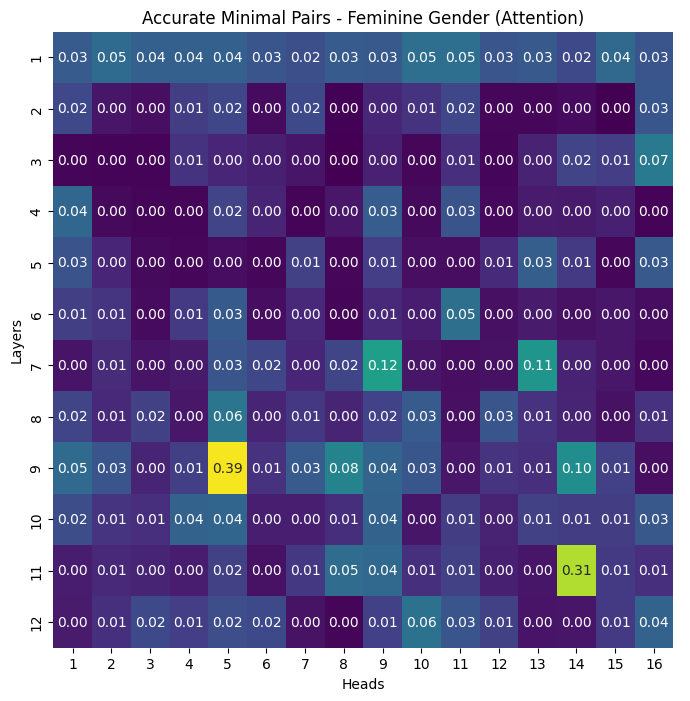}}
    \end{minipage}

    \vspace{2mm} 

    \begin{minipage}{0.45\textwidth}
        \centering
        \subfloat[mBART]{\includegraphics[width=\linewidth]{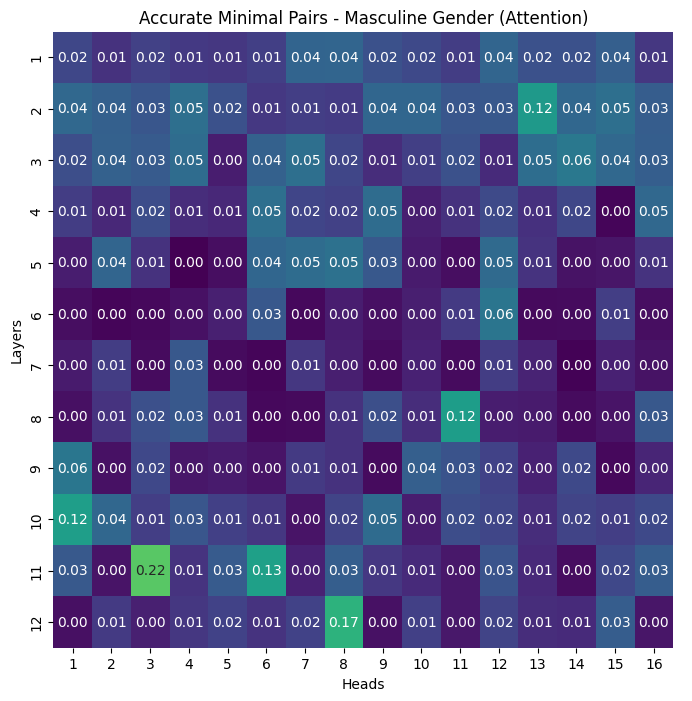}}
    \end{minipage}
    \hspace{-2mm}
    \begin{minipage}{0.45\textwidth}
        \centering
        \subfloat[mBART]{\includegraphics[width=\linewidth]{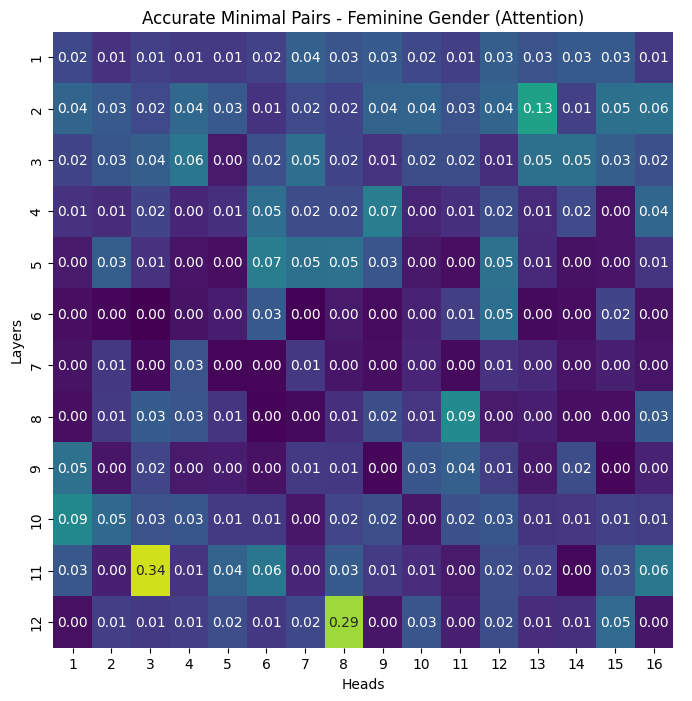}}
    \end{minipage}
    \vspace{2mm}
    \caption{\small{Heatmaps illustrating average encoder self-attention weights between the gender cue (\textit{i.e.}, pronoun) and the profession noun across accurate minimal pairs for each model. Each row contrasts masculine (left) \textit{vs.} feminine (right) referents, allowing for a comparison of how gender cues are integrated into target word representations. A standardized colormap is applied across all heatmaps.}}
    \label{fig:context_mix_2}
\end{figure*}

Given that the gender cue serves as the only explicit indicator of the primary entity’s gender in the source language (EN), it is expected to play a key role in the gender disambiguation of the target word in the target language (IT). To examine this, we focus on accurately gendered minimal pairs. We begin by identifying the target word’s source-side index by leveraging the annotations in WinoMT. We then align source and target sentences using \texttt{fast\_align}\footnote{\href{https://github.com/clab/fast_align}{github.com/clab/fast\_align}}  to retrieve the target word's corresponding index in the generated translation. The gender cue is identified by detecting a predefined set of pronouns (\textit{he, she, him, her, his}) in the source sentence, from which we extract their corresponding index. Since both target word and gender cue may be tokenized into multiple subwords, we map them accordingly by iterating through the tokenized sequence, incrementally matching subword segments. 
Once the relevant (subword) indices are obtained, we extract the corresponding attention weights from the model's attention matrices. To account subword tokenization, we compute the average attention weights across subword tokens before aggregating the values across all instances. Since attention weights sum to 1 across all context tokens, no further normalization is required.

\subsection{Results and Analysis}\label{att_analysis}

The heatmaps in Figure \ref{fig:context_mix_1} illustrate the self-attention weights between the gender cue and the target word, averaged across all sentences. These scores indicate how much the target word attends to the gender cue, \textit{i.e.}, the contribution of the cue to the target word’s contextualized representation. 


To identify attention heads that may play a more specialized role in gender disambiguation, we establish a threshold of relevance. Given that minimal pair sentences have an average length of $\approx13$ words, a uniform attention distribution would allocate a weight of approximately $1/13 \ (\approx0.08)$ to each word. Therefore, we consider attention heads that exceed this baseline by a notable margin as potentially relevant for gender cue integration. 

Comparing the models, we observe distinct attention patterns. While for OPUS\_MT a single attention head stands out at an early stage of encoding (layer 2), the other two models display a more distributed pattern, with at least two potentially influential attention heads emerging in deeper layers. This seems to indicate a more diffuse integration and a multi-layered processing of the gender cues. 

To further investigate whether models encode masculine and feminine gender cues differently, we separately report the attention scores for masculine and feminine pronouns in Figure \ref{fig:context_mix_2}. This reveals that feminine pronouns elicit more localized activations, while masculine ones tend to receive weaker, more dispersed attention, especially for OPUS-MT and mBART. Finally, NLLB-200 exhibits a different type of asymmetry, in which distinct attention heads appear to specialize in encoding gender-specific patterns -- some being more responsive to feminine pronouns, others playing a stronger role in encoding masculine ones.

While informative, these results must be interpreted with caution. As most feminine examples are found in pro-stereotypical settings (Table \ref{tab:gender_comp_mpa}), the observed attention patterns may reflect a form of training or dataset bias, where models have learned to associate certain professions with feminine pronouns due to their statistical distribution in the training data, rather than consistently relying on syntactic dependencies. Furthermore, combining this with the way minimal pairs are constructed, an inherent gender composition imbalance emerges. Since feminine entities are predominantly featured in pro-stereotypical examples, masculine ones are mostly found in anti-stereotypical settings. As a result, there are relatively fewer observations for pro-stereotypical masculine and anti-stereotypical feminine cases, making it difficult to draw definitive conclusions about gender cue integration in these underrepresented scenarios.

\section{Discussion}
In this section, we reflect on the key findings from our two-fold analysis, their implications, as well as potential avenues for future research. 

\subsection{Minimal Pair Accuracy and Default Masculinity}
While standard metrics of gender accuracy reveal that the investigated encoder-decoder models perform better for masculine referents and in pro-stereotypical settings, the proposed MPA uncovers another systematic asymmetries in gender disambiguation and exposes a persistent male-as-norm bias \citep{danesi-2014}. 

Although NLLB-200 and mBART showcase a more consistent integration of contextual information as compared to OPUS-MT, all models struggle to correctly disambiguate stereotypically male-dominated professions when provided with a feminine cue word while the reverse does not hold true. Namely, combining a stereotypically male profession with a feminine target cue (e.g., \textit{she}) often fails to trigger the corresponding feminine form, with models defaulting to the masculine variant. This asymmetry suggests a stronger bias towards masculine defaults, particularly in contexts where the feminine form challenges prevailing stereotypes. This asymmetry raises a more fundamental question of whether MT models can indeed consistently process syntactic dependencies for gender disambiguation or whether they are predominantly influenced by entrenched statistical associations. Our results seem to reinforce prior findings that language models often follow a default-to-masculine reasoning process when assigning gender \citep{jumelet-etal-2019-analysing, danesi-2014}, hence we wonder: \textit{Are we paying attention to \textbf{her}?}

\subsection{The Role of Attention in Gender Encoding}
As the model’s primary objective is translation, gender disambiguation is likely treated as an auxiliary task, with responsibility for its resolution distributed across various parts of the network, \textit{i.e.}, specific layers or attention heads within the model \citep{xu2015, wang2016, rocktaschel2016, lee-etal-2017-interactive, attention2017, clark-etal-2019-bert, kovaleva-etal-2019-revealing, reif2019, lin-etal-2019-open, voita-etal-2019-analyzing, jo-myaeng-2020-roles}. 

Having isolated accurate minimal pairs, we can speculate that the identified influential heads may specialize in encoding gender information during translation. Overall, these appear in early layers for OPUS-MT, mid-to-deep layers for NLLB-200, and deeper layers for mBART. Interestingly, gender cue integration is not uniform across all models and presents gender-specific patterns. Specifically, we observe that feminine pronouns elicit more localized activations, while masculine ones tend to receive weaker, more dispersed attention, especially for OPUS-MT and mBART. This aligns with prior research on gender representation in language models, which has shown that masculinity tends to function as the default category, while gender-specific signals -- particularly feminine ones -- are processed in a more localized manner \citep{jumelet-etal-2019-analysing, birth_of_bias}. Notably, NLLB-200 exhibits a different type of asymmetry, where distinct attention heads appear to specialize in encoding gender-specific patterns -- some being more responsive to feminine pronouns, others playing a stronger role in encoding masculine ones.

Expanding on these results, we find that models with more distributed and diffuse attention activation -- such as mBART and NLLB-200 -- perform better in terms of both gender accuracy and MPA compared to OPUS-MT, which attends gender cues in a single early-layer attention head. This suggests that gender disambiguation may benefit from a more adaptable, multi-layered gender encoding mechanism rather than a rigid, localized one. 


\subsection{Limitations and Future Work}
Our findings suggest potential avenues for binary gender bias mitigation strategies. Given that potentially influential attention heads have been identified, targeted interventions could be explored to enhance gender cue integration. Specifically, two promising directions include (i) fine-tuning seemingly specialized attention heads or (ii) enforcing a minimum attention threshold to ensure that gender cues receive sufficient weight when generating target words. 

Context mixing scores -- such as attention weights -- 
provide useful insights into how models may be processing gender cues and encoding gender-related information, especially when combined with nuanced evaluation metrics such as MPA. However, they should not be taken as definitive explanations of model decision-making as no causal relationship between gender cue integration and translation outputs is established. Although subsetting on accurately gendered minimal pairs partially addresses this limitation, it also introduces additional challenges. As discussed in Section \ref{att_analysis}, the gender composition imbalance within minimal pairs makes it difficult to assess whether observed attention patterns genuinely reflect contextual gender disambiguation or are simply a byproduct of learned statistical associations in the data. To address these challenges, future work should explore mechanistic interpretability methods -- such as activation patching \citep{vig2020cma, meng2022, heimersheim2024} -- to directly assess the causal role of gender cues in translation decisions.

\section{Conclusion}
In this work, we examined how Transformer-based NMT models integrate contextual gender cues and uncovered systematic biases and asymmetries in their processing mechanisms. 



Taken together, our findings reinforce previous calls for greater caution when interpreting benchmark scores for gender accuracy in MT \citep{savoldi2021gender}. Surface-level improvements, such as higher gender accuracy, can still obscure deeper biases in how and under which conditions these forms indeed appear. More nuanced and comprehensive analyses are needed to  to determine whether current systems truly leverage gender-specific cues or merely reinforce statistical stereotypes in subtler ways. 

Without a more careful consideration of when, why and how certain patterns emerge, we risk misinterpreting progress and overlooking specific persistent and more structural biases in MT. Ultimately, understanding how gender is encoded in translation models is a crucial component to ensure more fairness, accountability, and transparency in AI systems.

\section*{Acknowledgements}

We thank the reviewers for their insightful comments and feedback. We further extend our gratitude to our colleague Hosein Mohebbi for his critical suggestions and guidance, which helped shape the direction of this work.

\bibliography{mtsummit25, anthology}

\appendix
\section{Cross-Attention Analysis}
\label{sec:appendix_cross_attention}

In this section, we present the average cross-attention weights, illustrating how the decoder attends to the gender cue (\textit{i.e.}, the pronoun) in the encoder representations when generating the target word (\textit{i.e.}, the gendered profession).  

\begin{figure}[htbp]
    \centering

    \begin{minipage}{0.45\textwidth}
        \centering
        \subfloat[OPUS-MT]{\includegraphics[width=\linewidth]{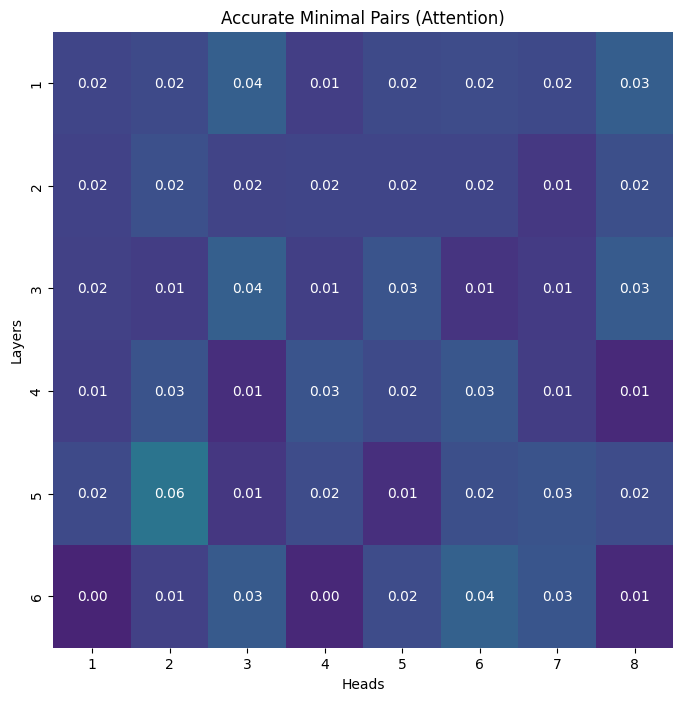}} 
    \end{minipage}

    \vspace{2mm} 

    \begin{minipage}{0.45\textwidth}
        \centering
        \subfloat[NLLB-200]{\includegraphics[width=\linewidth]{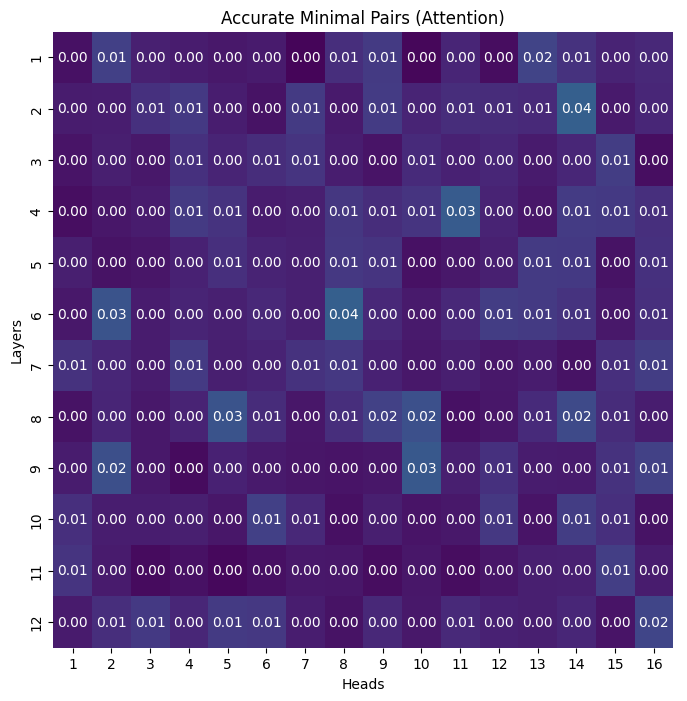}} 
    \end{minipage}

    \vspace{2mm} 

    \begin{minipage}{0.45\textwidth}
        \centering
        \subfloat[mBART]{\includegraphics[width=\linewidth]{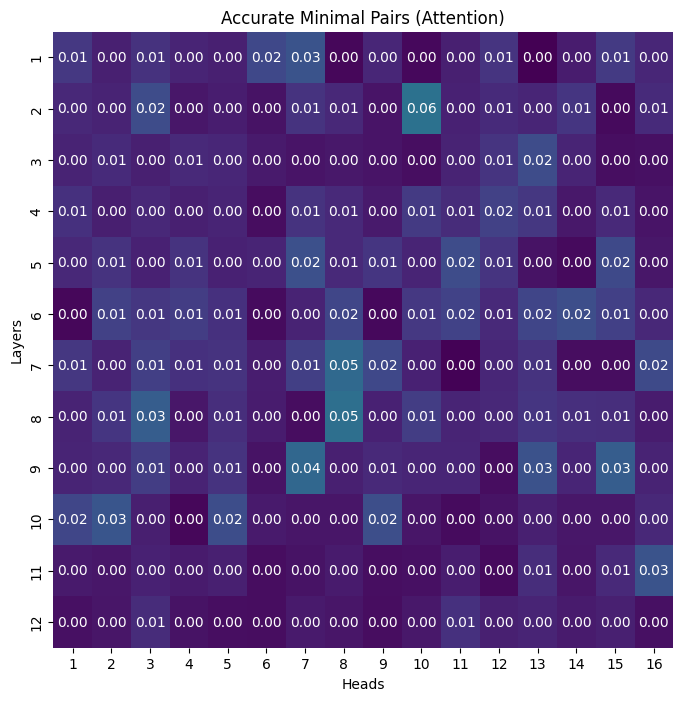}} 
    \end{minipage}
    
    \vspace{2mm}
     
    \caption{\small{Heatmaps illustrating average cross-attention weights to the gender cue (\textit{i.e.}, pronoun) when generating the profession noun across accurate minimal pairs for each model. A standardized colormap is applied across all heatmaps.}} 
    \label{fig:cross_attention_1}
\end{figure}

\begin{figure*}[htbp]
    \centering
    \begin{minipage}{0.45\textwidth}
        \centering
        \subfloat[OPUS-MT]{\includegraphics[width=\linewidth]{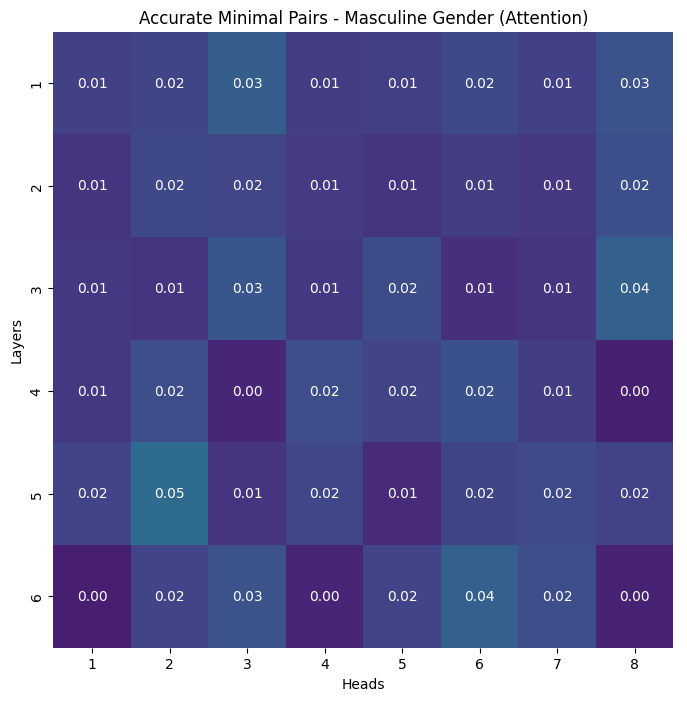}} 
    \end{minipage}
    \hspace{-2mm}
    \begin{minipage}{0.45\textwidth}
        \centering
        \subfloat[OPUS-MT]{\includegraphics[width=\linewidth]{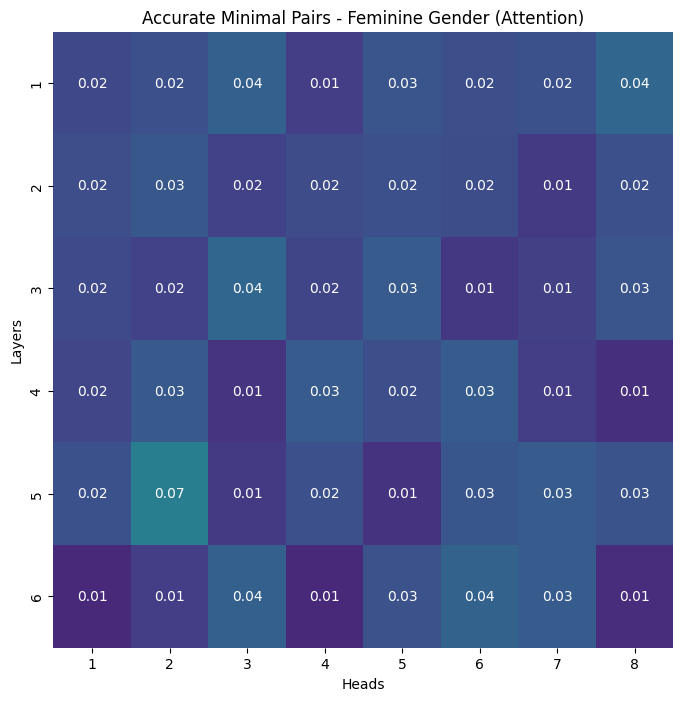}} 
    \end{minipage}

    \vspace{2mm} 
    
    \begin{minipage}{0.45\textwidth}
        \centering
        \subfloat[NLLB-200]{\includegraphics[width=\linewidth]{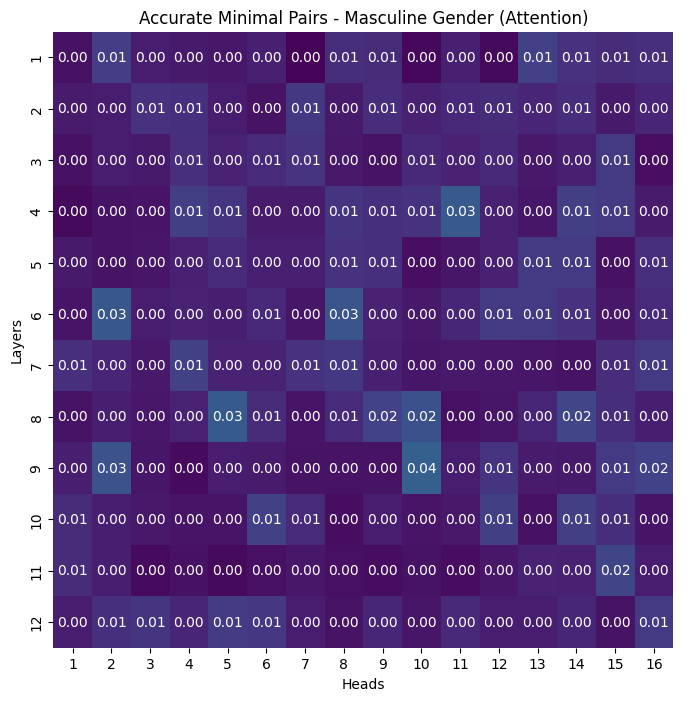}} 
    \end{minipage}
    \hspace{-2mm}
    \begin{minipage}{0.45\textwidth}
        \centering
        \subfloat[NLLB-200]{\includegraphics[width=\linewidth]{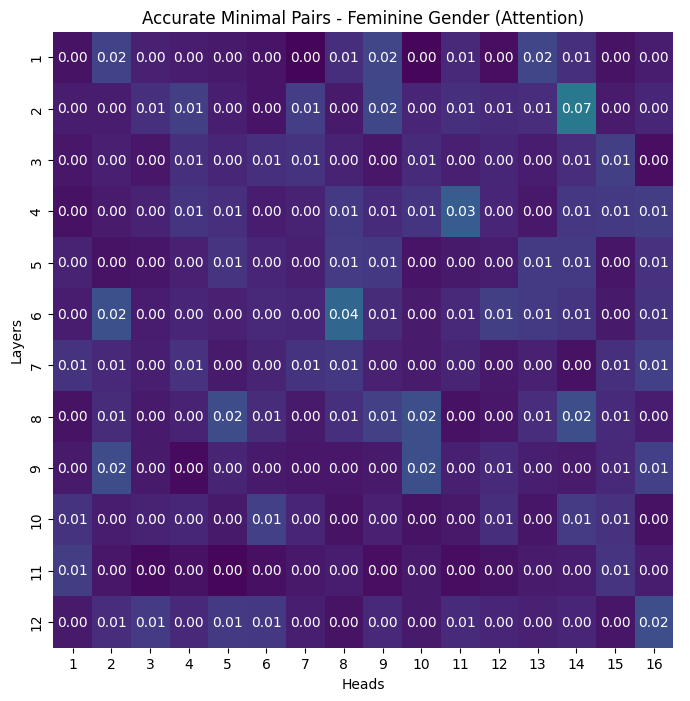}} 
    \end{minipage}

    \vspace{2mm} 

    \begin{minipage}{0.45\textwidth}
        \centering
        \subfloat[mBART]{\includegraphics[width=\linewidth]{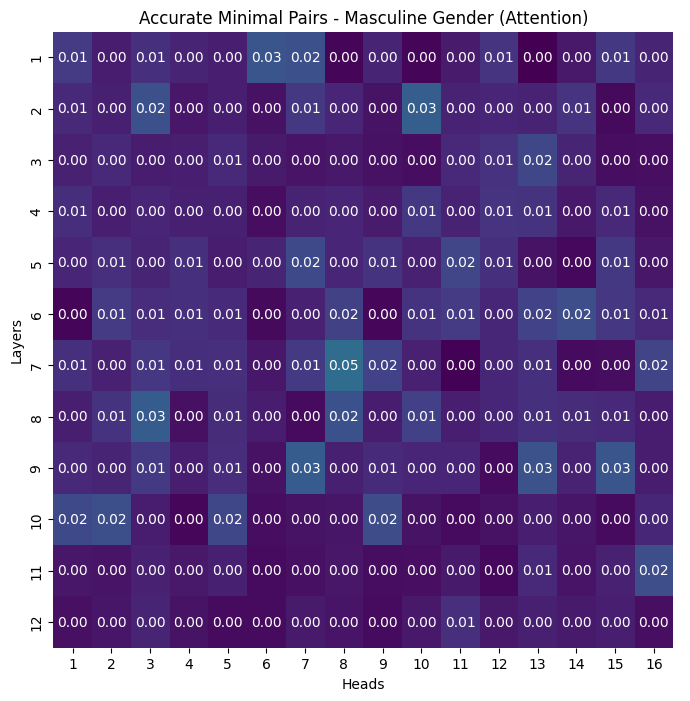}} 
    \end{minipage}
    \hspace{-2mm}
    \begin{minipage}{0.45\textwidth}
        \centering
        \subfloat[mBART]{\includegraphics[width=\linewidth]{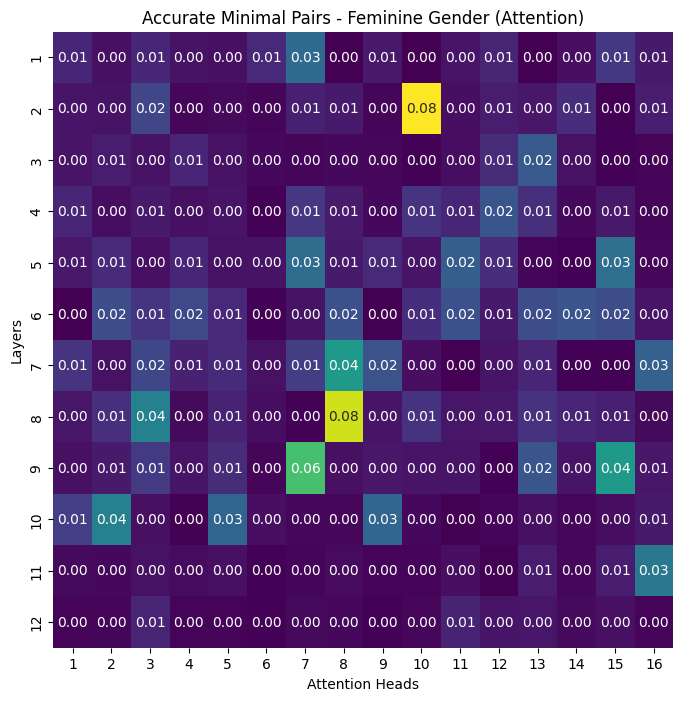}} 
    \end{minipage}
    
    \vspace{2mm}
    
    \caption{\small{Heatmaps illustrating average cross-attention weights to the gender cue (\textit{i.e.}, pronoun) when generating the profession noun across accurate minimal pairs for each model. Each row contrasts masculine (left) \textit{vs.} feminine (right) referents, allowing for a comparison of how gender cues are integrated into target word representations at the decoder level. A standardized colormap is applied across all heatmaps.}} 
    \label{fig:cross_attention_2}
\end{figure*}


\end{document}

%% file: notes/intro_EVA.tex
The field of Machine Translation (MT) has undergone significant technological shifts over the past decades, moving from transparent rule-based systems to increasingly opaque probability-based ones such as statistical and neural MT. Furthermore, the complexity and scale of current Transformer-based \citep{attention2017} architectures, which underpin both neural MT (NMT) and Large Language Models (LLMs), are making it more challenging to trace back model decisions and understand the underlying processes. This growing opacity raises concerns for AI governance where transparency, fairness and risk mitigation are becoming increasingly important for a responsible deployment of MT technology.



At the same time, research on (gender) bias in MT has been on the rise, reflecting more general tendencies in the field of Natural Language Processing (NLP) \citep{sun-etal-2019-mitigating, costa2019analysis,blodgett-etal-2020-language, stanczak2021survey}. 
The increasing awareness has led to concerns related to the flaws, inconsistencies and biases that models inherit, propagate and potentially exacerbate -- especially with the increasing integration of NLP tools in people's everyday lives \citep{bansal2022survey}. 
In response, AI governance policies are emerging worldwide, such as the European Union’s AI Act (\citeyear{EU_AI_ACT_2024}), aiming to regulate the development and deployment of AI systems to ensure ethical standards and mitigate potential risks. 
For MT specifically, the nature of the translation task itself further complicates matters due to cross-linguistic differences in gender representation and expression across languages, where social gender, linguistic gender and diverse cultural contexts intersect.
\vspace{1em}
\begin{figure}[h!]
    \centering

    \begin{tcolorbox}[
        colback=blue!4, colframe=blue!50, 
        boxrule=0.8pt, arc=8pt, 
        left=6pt, right=6pt, top=4pt, bottom=4pt 
    ]
        \textbf{EN:} \hspace{0.5em} The cook prepared a soup for the \textcolor{purple}{\textbf{housekeeper}} because \textcolor{purple}{\textbf{he}} helped clean the room.
    \end{tcolorbox}
    \begin{tcolorbox}[
        colback=purple!4, colframe=purple!50, 
        boxrule=0.8pt, arc=8pt,
        left=6pt, right=6pt, top=4pt, bottom=4pt 
    ]
        \textbf{IT:} \hspace{0.5em} Il cuoco ha preparato una zuppa per \textcolor{orange}{\textbf{la governante}} perché ha aiutato a pulire la stanza.
    \end{tcolorbox}

    \caption[]{\small{Example from the WinoMT dataset~\citep{winomt} illustrating gender bias in an English-Italian translation. While the English sentence establishes the referent as male (using the pronoun \textit{he}), the translation\footnotemark ~uses a feminine form \textit{la governante}, thereby disregarding the contextual gender cue.}\label{fig_1:gender_bias_example}}
\end{figure}
\footnotetext{Generated by ChatGPT on March 6th, 2025.} 

\newpage
Languages encode gender in different ways and to varying degrees \citep{ackerman, cao-daume}. While some, such as English or Danish, rely predominantly on pronouns, others, such as Italian, require morphological agreement across multiple parts of speech \citep{stahlberg}. This implies that -- in certain translation contexts -- implicit source information must be made explicit in the target \citep{vanmassenhove-etal-2018-getting}. Figure~\ref{fig_1:gender_bias_example} illustrates this through an example from the WinoMT dataset \citep{winomt}. The English word  \textit{housekeeper} is translated into the Italian feminine form \textit{la governante}, despite the broader sentence context indicating that the referent identifies as male (\textit{he}). When this happens on a large scale and in a systematic way, it can result in representational and allocational harms, disproportionately affecting more marginalized groups \citep{blodgett-etal-2020-language} while simultaneously eroding linguistic diversity \citep{vanmassenhove2019lost,mtese}.

Despite the increasing awareness and research efforts over the past decade \citep{review_gender_bias_mt}, gender bias in MT remains a complex, largely unsolved challenge \citep{llm_bias_4, gender_bias_llms2}. While current evaluation metrics offer a broad bias assessment, they do not capture whether models actively integrate contextual cues or default to learned statistical associations when disambiguating gendered nouns. This limitation makes it challenging to determine whether observed errors stem from a failure to process contextual information, the reinforcement of pre-existing biases, or internal shortcomings in how gender information is encoded and utilized. This hinders the development of targeted interventions and effective mitigation strategies.
\vspace{1em}

To address this gap, we provide a nuanced evaluation framework that moves beyond a surface-level assessment of gender realization in an English-Italian translation context. Our main contribution is two-fold:





\begin{itemize}
    \item We introduce \textbf{Minimal Pair Accuracy (MPA)}, a novel metric that measures whether models consistently rely on gender cues for gender disambiguation, rather than defaulting to learned priors. By leveraging the WinoMT dataset \citep{winomt}, we construct minimal pairs, \textit{i.e.}, sentence pairs that only differ in the gendered pronoun, and compute the proportion of cases where the model correctly adjusts the target gender. 
    \item We conduct an exploratory \textbf{Attention-Based Analysis} to better understand how gender information is encoded within Transformer models. Specifically, we examine the extent to which profession nouns attend to gender cues at different layers and attention heads, and whether this behavior varies based on gender (masculine \textit{vs.} feminine) or alignment with gender-role stereotypes (pro-stereotypical \textit{vs.} anti-stereotypical contexts). 
\end{itemize}

%% file: notes/related_work_EVA.tex

Research on gender bias in MT has largely focused on: analyzing MT output (e.g. \citet{rescigno2020case,ramesh2021evaluating}...); rewriting into gendered (e.g. \citet{vanmassenhove-etal-2018-getting,moryossef2019filling, habash2019automatic} or neutral outputs (e.g. \citet{vanmassenhove2021neutral,sun2021they}...); word-embedding debiasing techniques (e.g. \citet{hirasawa2019debiasing,font2019equalizing}....), domain adaptation (e.g. \citet{saunders2020reducing}), counterfactual data augmentation (e.g. \citet{zmigrod2019counterfactual}) and/or the development novel benchmarks and evaluation sets (e.g. \citet{winomt,luisa2020gender}....). Given that several studies \citep{blodgett-etal-2020-language, stanczak2021survey,savoldi2021gender} already offer a more comprehensive overview of broader discussions and research on (gender) bias in language technology, we specifically dedicate this related work section to the limited body of work focusing on the internal mechanisms underlying gender bias in (MT) models and interpretability techniques.

MT-specific research on interpretability techniques has largely focused on linguistic competence through probing \citep{belinkov-etal-2017-neural,belinkov-etal-2017-evaluating, conneau-etal-2018-cram}, or by analyzing contrastive translation \citep{sennrich-2017-grammatical, burlot-yvon-2017-evaluating, rios-gonzales-etal-2017-improving, vamvas-sennrich-2021-contrastive, vamvas-sennrich-2022-little}. More recent work investigated how MT systems process intra- and inter-sentential context and whether their context usage aligns with human expectations \citep{goindani-shrivastava-2021-dynamic, voita-etal-2021-analyzing, sarti2024_context, mohammed-niculae-2024-measuring}. Despite high overall performance, these studies highlight how models often struggle to effectively leverage contextual information, either failing to integrate necessary information or attending to irrelevant tokens when resolving ambiguities \citep{kim-etal-2019-document, yin-etal-2021-context}, an interesting finding raising concerns about gender disambiguation which indeed could be driven by biased statistical patterns rather than reliance on relevant contextual cues.

The problem of context integration is not only relevant to model decision-making but also affects how gender bias is evaluated. Template-based evaluation frameworks, such as WinoMT \citep{winomt}, provide controlled settings to measure surface-level accuracy metrics, and have been widely used to quantify gender bias across different language pairs and MT systems \citep{kocmi-etal-2020-gender, costajussa2020, choubey-etal-2021-gfst}. However, as these primarily rely on the alignment and morphosyntactic analysis of lexically gender-ambiguous words, they do not reveal whether models actively integrate contextual cues when making gender-related decisions. These limitations underscore the need for more nuanced evaluation methods.

A promising avenue for investigating how gender cues influence model decisions is through the study of context mixing, \textit{i.e.}, the ability of Transformer-based models to dynamically incorporate information from the broader context into token representations. This process is largely governed by the attention mechanism, which plays a central role in these models. While attention-based analyses  have been criticized for their reliability \citep{jain-wallace-2019-attention, bibal-etal-2022-attention}, and more advanced interpretability methods have been introduced \citep{kobayashi-etal-2020-attention, kobayashi-etal-2021-incorporating, modarressi-etal-2022-globenc, ferrando-etal-2022-measuring, mohebbi-etal-2023-quantifying}, attention weights remain a popular choice for analyzing model behavior due to their ability to provide direct insights into token interactions across layers and heads. As a matter of fact, they have been extensively leveraged to track token dependencies, revealing that specific attention heads may specialize in distinct linguistic functions \citep{xu2015,rocktaschel2016, wang2016, lee-etal-2017-interactive, attention2017, kovaleva-etal-2019-revealing, reif2019, lin-etal-2019-open, voita-etal-2019-analyzing, jo-myaeng-2020-roles}.

To the best of our knowledge, only the study by \citet{bau-2018} attempted to control gender through internal mechanisms in an MT setting. They explored this by probing and deactivating specific neurons associated with gender in an Long Short-Term Memory (LSTM) architecture. Their findings showed that gender-related properties are widely distributed across the network, making effectively controlling the output very difficult.

%% file: mtsummit25.bbl
\begin{thebibliography}{77}
\providecommand{\natexlab}[1]{#1}

\bibitem[{Ackerman(2019)}]{ackerman}
Lauren Ackerman. 2019.
\newblock \href {https://doi.org/10.5334/gjgl.721} {Syntactic and cognitive issues in investigating gendered coreference}.
\newblock \emph{Glossa: a journal of general linguistics}, 4(1).

\bibitem[{Bansal(2022)}]{bansal2022survey}
Rajas Bansal. 2022.
\newblock A survey on bias and fairness in natural language processing.
\newblock \emph{arXiv preprint arXiv:2204.09591}.

\bibitem[{Bau et~al.(2018)Bau, Belinkov, Sajjad, Durrani, Dalvi, and Glass}]{bau-2018}
Anthony Bau, Yonatan Belinkov, Hassan Sajjad, Nadir Durrani, Fahim Dalvi, and James Glass. 2018.
\newblock \href {https://arxiv.org/abs/1811.01157} {Identifying and controlling important neurons in neural machine translation}.
\newblock In \emph{Proceedings of the Seventh International Conference on Learning Representations (ICLR)}.

\bibitem[{Belinkov et~al.(2017{\natexlab{a}})Belinkov, Durrani, Dalvi, Sajjad, and Glass}]{belinkov-etal-2017-neural}
Yonatan Belinkov, Nadir Durrani, Fahim Dalvi, Hassan Sajjad, and James Glass. 2017{\natexlab{a}}.
\newblock \href {https://doi.org/10.18653/v1/P17-1080} {What do neural machine translation models learn about morphology?}
\newblock In \emph{Proceedings of the 55th Annual Meeting of the Association for Computational Linguistics (Volume 1: Long Papers)}, pages 861--872, Vancouver, Canada. Association for Computational Linguistics.

\bibitem[{Belinkov et~al.(2017{\natexlab{b}})Belinkov, M{\`a}rquez, Sajjad, Durrani, Dalvi, and Glass}]{belinkov-etal-2017-evaluating}
Yonatan Belinkov, Llu{\'i}s M{\`a}rquez, Hassan Sajjad, Nadir Durrani, Fahim Dalvi, and James Glass. 2017{\natexlab{b}}.
\newblock \href {https://aclanthology.org/I17-1001/} {Evaluating layers of representation in neural machine translation on part-of-speech and semantic tagging tasks}.
\newblock In \emph{Proceedings of the Eighth International Joint Conference on Natural Language Processing (Volume 1: Long Papers)}, pages 1--10, Taipei, Taiwan. Asian Federation of Natural Language Processing.

\bibitem[{Bibal et~al.(2022)Bibal, Cardon, Alfter, Wilkens, Wang, Fran{\c{c}}ois, and Watrin}]{bibal-etal-2022-attention}
Adrien Bibal, R{\'e}mi Cardon, David Alfter, Rodrigo Wilkens, Xiaoou Wang, Thomas Fran{\c{c}}ois, and Patrick Watrin. 2022.
\newblock \href {https://doi.org/10.18653/v1/2022.acl-long.269} {Is attention explanation? an introduction to the debate}.
\newblock In \emph{Proceedings of the 60th Annual Meeting of the Association for Computational Linguistics (Volume 1: Long Papers)}, pages 3889--3900, Dublin, Ireland. Association for Computational Linguistics.

\bibitem[{Blodgett et~al.(2020)Blodgett, Barocas, Daum{\'e}~III, and Wallach}]{blodgett-etal-2020-language}
Su~Lin Blodgett, Solon Barocas, Hal Daum{\'e}~III, and Hanna Wallach. 2020.
\newblock \href {https://doi.org/10.18653/v1/2020.acl-main.485} {Language (technology) is power: A critical survey of {\textquotedblleft}bias{\textquotedblright} in {NLP}}.
\newblock In \emph{Proceedings of the 58th Annual Meeting of the Association for Computational Linguistics}, pages 5454--5476, Online. Association for Computational Linguistics.

\bibitem[{Burlot and Yvon(2017)}]{burlot-yvon-2017-evaluating}
Franck Burlot and Fran{\c{c}}ois Yvon. 2017.
\newblock \href {https://doi.org/10.18653/v1/W17-4705} {Evaluating the morphological competence of machine translation systems}.
\newblock In \emph{Proceedings of the Second Conference on Machine Translation}, pages 43--55, Copenhagen, Denmark. Association for Computational Linguistics.

\bibitem[{Cao and Daum{\'e}~III(2020)}]{cao-daume}
Yang~Trista Cao and Hal Daum{\'e}~III. 2020.
\newblock \href {https://doi.org/10.18653/v1/2020.acl-main.418} {Toward gender-inclusive coreference resolution}.
\newblock In \emph{Proceedings of the 58th Annual Meeting of the Association for Computational Linguistics}, pages 4568--4595. Association for Computational Linguistics.

\bibitem[{Choubey et~al.(2021)Choubey, Currey, Mathur, and Dinu}]{choubey-etal-2021-gfst}
Prafulla~Kumar Choubey, Anna Currey, Prashant Mathur, and Georgiana Dinu. 2021.
\newblock \href {https://doi.org/10.18653/v1/2021.emnlp-main.123} {{GFST}: {G}ender-filtered self-training for more accurate gender in translation}.
\newblock In \emph{Proceedings of the 2021 Conference on Empirical Methods in Natural Language Processing}, pages 1640--1654, Online and Punta Cana, Dominican Republic. Association for Computational Linguistics.

\bibitem[{Clark et~al.(2019)Clark, Khandelwal, Levy, and Manning}]{clark-etal-2019-bert}
Kevin Clark, Urvashi Khandelwal, Omer Levy, and Christopher~D. Manning. 2019.
\newblock \href {https://doi.org/10.18653/v1/W19-4828} {What does {BERT} look at? an analysis of {BERT}`s attention}.
\newblock In \emph{Proceedings of the 2019 ACL Workshop BlackboxNLP: Analyzing and Interpreting Neural Networks for NLP}, pages 276--286, Florence, Italy. Association for Computational Linguistics.

\bibitem[{Conneau et~al.(2018)Conneau, Kruszewski, Lample, Barrault, and Baroni}]{conneau-etal-2018-cram}
Alexis Conneau, German Kruszewski, Guillaume Lample, Lo{\"i}c Barrault, and Marco Baroni. 2018.
\newblock \href {https://doi.org/10.18653/v1/P18-1198} {What you can cram into a single {\$}{\&}!{\#}* vector: Probing sentence embeddings for linguistic properties}.
\newblock In \emph{Proceedings of the 56th Annual Meeting of the Association for Computational Linguistics (Volume 1: Long Papers)}, pages 2126--2136, Melbourne, Australia. Association for Computational Linguistics.

\bibitem[{Costa-juss{\`a}(2019)}]{costa2019analysis}
Marta~R Costa-juss{\`a}. 2019.
\newblock {An analysis of Gender Bias studies in Natural Language Processing}.
\newblock \emph{Nature Machine Intelligence}, 1(11):495--496.

\bibitem[{Costa-jussà et~al.(2022)Costa-jussà, Cross, Çelebi, Elbayad, Heafield, Heffernan, Kalbassi, Lam, Licht, Maillard, Sun, Wang, Wenzek, Youngblood, Akula, Barrault, Gonzalez, Hansanti, Hoffman, Jarrett, Sadagopan, Rowe, Spruit, Tran, Andrews, Ayan, Bhosale, Edunov, Fan, Gao, Goswami, Guzmán, Koehn, Mourachko, Ropers, Saleem, Schwenk, and Wang}]{nllb200}
Marta~R. Costa-jussà, James Cross, Onur Çelebi, Maha Elbayad, Kenneth Heafield, Kevin Heffernan, Elahe Kalbassi, Janice Lam, Daniel Licht, Jean Maillard, Anna Sun, Skyler Wang, Guillaume Wenzek, Al~Youngblood, Bapi Akula, Loic Barrault, Gabriel~Mejia Gonzalez, Prangthip Hansanti, John Hoffman, Semarley Jarrett, Kaushik~Ram Sadagopan, Dirk Rowe, Shannon Spruit, Chau Tran, Pierre Andrews, Necip~Fazil Ayan, Shruti Bhosale, Sergey Edunov, Angela Fan, Cynthia Gao, Vedanuj Goswami, Francisco Guzmán, Philipp Koehn, Alexandre Mourachko, Christophe Ropers, Safiyyah Saleem, Holger Schwenk, and Jeff Wang. 2022.
\newblock \href {https://arxiv.org/abs/2207.04672} {No language left behind: Scaling human-centered machine translation}.
\newblock \emph{Preprint}, arXiv:2207.04672.

\bibitem[{Costa-jussà et~al.(2020)Costa-jussà, Escolano, Basta, Ferrando, Batlle, and Kharitonova}]{costajussa2020}
Marta~R. Costa-jussà, Carlos Escolano, Christine Basta, Javier Ferrando, Roser Batlle, and Ksenia Kharitonova. 2020.
\newblock \href {https://arxiv.org/abs/2012.13176} {Gender bias in multilingual neural machine translation: The architecture matters}.
\newblock \emph{Preprint}, arXiv:2012.13176.

\bibitem[{Danesi(2014)}]{danesi-2014}
Marcel Danesi. 2014.
\newblock \href {https://doi.org/10.4324/9781315705194} {\emph{{Dictionary of Media and Communications}}}.
\newblock Routledge.

\bibitem[{Ferrando et~al.(2022)Ferrando, G{\'a}llego, and Costa-juss{\`a}}]{ferrando-etal-2022-measuring}
Javier Ferrando, Gerard~I. G{\'a}llego, and Marta~R. Costa-juss{\`a}. 2022.
\newblock \href {https://doi.org/10.18653/v1/2022.emnlp-main.595} {Measuring the mixing of contextual information in the transformer}.
\newblock In \emph{Proceedings of the 2022 Conference on Empirical Methods in Natural Language Processing}, pages 8698--8714, Abu Dhabi, United Arab Emirates. Association for Computational Linguistics.

\bibitem[{Font and Costa-juss{\`a}(2019)}]{font2019equalizing}
Joel~Escud{\'e} Font and Marta~R Costa-juss{\`a}. 2019.
\newblock Equalizing gender bias in neural machine translation with word embeddings techniques.
\newblock In \emph{Proceedings of the First Workshop on Gender Bias in Natural Language Processing}, pages 147--154.

\bibitem[{Goindani and Shrivastava(2021)}]{goindani-shrivastava-2021-dynamic}
Akshay Goindani and Manish Shrivastava. 2021.
\newblock \href {https://aclanthology.org/2021.ranlp-1.52/} {A dynamic head importance computation mechanism for neural machine translation}.
\newblock In \emph{Proceedings of the International Conference on Recent Advances in Natural Language Processing (RANLP 2021)}, pages 454--462, Held Online. INCOMA Ltd.

\bibitem[{Habash et~al.(2019)Habash, Bouamor, and Chung}]{habash2019automatic}
Nizar Habash, Houda Bouamor, and Christine Chung. 2019.
\newblock Automatic gender identification and reinflection in arabic.
\newblock In \emph{Proceedings of the First Workshop on Gender Bias in Natural Language Processing}, pages 155--165.

\bibitem[{Heimersheim and Nanda(2024)}]{heimersheim2024}
Stefan Heimersheim and Neel Nanda. 2024.
\newblock \href {https://arxiv.org/abs/2404.15255} {How to use and interpret activation patching}.
\newblock \emph{Preprint}, arXiv:2404.15255.

\bibitem[{Hirasawa and Komachi(2019)}]{hirasawa2019debiasing}
Tosho Hirasawa and Mamoru Komachi. 2019.
\newblock Debiasing word embeddings improves multimodal machine translation.
\newblock In \emph{Proceedings of Machine Translation Summit XVII: Research Track}, pages 32--42.

\bibitem[{Jain and Wallace(2019)}]{jain-wallace-2019-attention}
Sarthak Jain and Byron~C. Wallace. 2019.
\newblock \href {https://doi.org/10.18653/v1/N19-1357} {{A}ttention is not {E}xplanation}.
\newblock In \emph{Proceedings of the 2019 Conference of the North {A}merican Chapter of the Association for Computational Linguistics: Human Language Technologies, Volume 1 (Long and Short Papers)}, pages 3543--3556, Minneapolis, Minnesota. Association for Computational Linguistics.

\bibitem[{Jo and Myaeng(2020)}]{jo-myaeng-2020-roles}
Jae-young Jo and Sung-Hyon Myaeng. 2020.
\newblock \href {https://doi.org/10.18653/v1/2020.acl-main.311} {Roles and utilization of attention heads in transformer-based neural language models}.
\newblock In \emph{Proceedings of the 58th Annual Meeting of the Association for Computational Linguistics}, pages 3404--3417, Online. Association for Computational Linguistics.

\bibitem[{Jumelet et~al.(2019)Jumelet, Zuidema, and Hupkes}]{jumelet-etal-2019-analysing}
Jaap Jumelet, Willem Zuidema, and Dieuwke Hupkes. 2019.
\newblock \href {https://doi.org/10.18653/v1/K19-1001} {Analysing neural language models: Contextual decomposition reveals default reasoning in number and gender assignment}.
\newblock In \emph{Proceedings of the 23rd Conference on Computational Natural Language Learning (CoNLL)}, pages 1--11, Hong Kong, China. Association for Computational Linguistics.

\bibitem[{Kim et~al.(2019)Kim, Tran, and Ney}]{kim-etal-2019-document}
Yunsu Kim, Duc~Thanh Tran, and Hermann Ney. 2019.
\newblock \href {https://doi.org/10.18653/v1/D19-6503} {When and why is document-level context useful in neural machine translation?}
\newblock In \emph{Proceedings of the Fourth Workshop on Discourse in Machine Translation (DiscoMT 2019)}, pages 24--34, Hong Kong, China. Association for Computational Linguistics.

\bibitem[{Kobayashi et~al.(2020)Kobayashi, Kuribayashi, Yokoi, and Inui}]{kobayashi-etal-2020-attention}
Goro Kobayashi, Tatsuki Kuribayashi, Sho Yokoi, and Kentaro Inui. 2020.
\newblock \href {https://doi.org/10.18653/v1/2020.emnlp-main.574} {Attention is not only a weight: Analyzing transformers with vector norms}.
\newblock In \emph{Proceedings of the 2020 Conference on Empirical Methods in Natural Language Processing (EMNLP)}, pages 7057--7075, Online. Association for Computational Linguistics.

\bibitem[{Kobayashi et~al.(2021)Kobayashi, Kuribayashi, Yokoi, and Inui}]{kobayashi-etal-2021-incorporating}
Goro Kobayashi, Tatsuki Kuribayashi, Sho Yokoi, and Kentaro Inui. 2021.
\newblock \href {https://doi.org/10.18653/v1/2021.emnlp-main.373} {{I}ncorporating {R}esidual and {N}ormalization {L}ayers into {A}nalysis of {M}asked {L}anguage {M}odels}.
\newblock In \emph{Proceedings of the 2021 Conference on Empirical Methods in Natural Language Processing}, pages 4547--4568, Online and Punta Cana, Dominican Republic. Association for Computational Linguistics.

\bibitem[{Kocmi et~al.(2020)Kocmi, Limisiewicz, and Stanovsky}]{kocmi-etal-2020-gender}
Tom Kocmi, Tomasz Limisiewicz, and Gabriel Stanovsky. 2020.
\newblock \href {https://aclanthology.org/2020.wmt-1.39/} {Gender coreference and bias evaluation at {WMT} 2020}.
\newblock In \emph{Proceedings of the Fifth Conference on Machine Translation}, pages 357--364, Online. Association for Computational Linguistics.

\bibitem[{Kovaleva et~al.(2019)Kovaleva, Romanov, Rogers, and Rumshisky}]{kovaleva-etal-2019-revealing}
Olga Kovaleva, Alexey Romanov, Anna Rogers, and Anna Rumshisky. 2019.
\newblock \href {https://doi.org/10.18653/v1/D19-1445} {Revealing the dark secrets of {BERT}}.
\newblock In \emph{Proceedings of the 2019 Conference on Empirical Methods in Natural Language Processing and the 9th International Joint Conference on Natural Language Processing (EMNLP-IJCNLP)}, pages 4365--4374, Hong Kong, China. Association for Computational Linguistics.

\bibitem[{Lee et~al.(2017)Lee, Shin, and Kim}]{lee-etal-2017-interactive}
Jaesong Lee, Joong-Hwi Shin, and Jun-Seok Kim. 2017.
\newblock \href {https://doi.org/10.18653/v1/D17-2021} {Interactive visualization and manipulation of attention-based neural machine translation}.
\newblock In \emph{Proceedings of the 2017 Conference on Empirical Methods in Natural Language Processing: System Demonstrations}, pages 121--126, Copenhagen, Denmark. Association for Computational Linguistics.

\bibitem[{Lin et~al.(2019)Lin, Tan, and Frank}]{lin-etal-2019-open}
Yongjie Lin, Yi~Chern Tan, and Robert Frank. 2019.
\newblock \href {https://doi.org/10.18653/v1/W19-4825} {Open sesame: Getting inside {BERT}`s linguistic knowledge}.
\newblock In \emph{Proceedings of the 2019 ACL Workshop BlackboxNLP: Analyzing and Interpreting Neural Networks for NLP}, pages 241--253, Florence, Italy. Association for Computational Linguistics.

\bibitem[{Liu et~al.(2020)Liu, Gu, Goyal, Li, Edunov, Ghazvininejad, Lewis, and Zettlemoyer}]{liu-etal-2020-multilingual-denoising}
Yinhan Liu, Jiatao Gu, Naman Goyal, Xian Li, Sergey Edunov, Marjan Ghazvininejad, Mike Lewis, and Luke Zettlemoyer. 2020.
\newblock \href {https://doi.org/10.1162/tacl_a_00343} {Multilingual denoising pre-training for neural machine translation}.
\newblock \emph{Transactions of the Association for Computational Linguistics}, 8:726--742.

\bibitem[{Luisa et~al.(2020)Luisa, Savoldi, Matteo, Di~Gangi~Mattia, Roldano, Marco et~al.}]{luisa2020gender}
Bentivogli Luisa, Beatrice Savoldi, Negri Matteo, A~Di~Gangi~Mattia, Cattoni Roldano, Turchi Marco, et~al. 2020.
\newblock Gender in danger? evaluating speech translation technology on the must-she corpus.
\newblock In \emph{Proceedings of the 58th Annual Meeting of the Association for Computational Linguistics}, pages 6923--6933. Association for Computational Linguistics.

\bibitem[{Meng et~al.(2022)Meng, Bau, Andonian, and Belinkov}]{meng2022}
Kevin Meng, David Bau, Alex Andonian, and Yonatan Belinkov. 2022.
\newblock \href {https://proceedings.neurips.cc/paper_files/paper/2022/file/6f1d43d5a82a37e89b0665b33bf3a182-Paper-Conference.pdf} {Locating and editing factual associations in gpt}.
\newblock In \emph{Advances in Neural Information Processing Systems}, volume~35, pages 17359--17372. Curran Associates, Inc.

\bibitem[{Modarressi et~al.(2022)Modarressi, Fayyaz, Yaghoobzadeh, and Pilehvar}]{modarressi-etal-2022-globenc}
Ali Modarressi, Mohsen Fayyaz, Yadollah Yaghoobzadeh, and Mohammad~Taher Pilehvar. 2022.
\newblock \href {https://doi.org/10.18653/v1/2022.naacl-main.19} {{G}lob{E}nc: Quantifying global token attribution by incorporating the whole encoder layer in transformers}.
\newblock In \emph{Proceedings of the 2022 Conference of the North American Chapter of the Association for Computational Linguistics: Human Language Technologies}, pages 258--271, Seattle, United States. Association for Computational Linguistics.

\bibitem[{Mohammed and Niculae(2024)}]{mohammed-niculae-2024-measuring}
Wafaa Mohammed and Vlad Niculae. 2024.
\newblock \href {https://aclanthology.org/2024.findings-eacl.113/} {On measuring context utilization in document-level {MT} systems}.
\newblock In \emph{Findings of the Association for Computational Linguistics: EACL 2024}, pages 1633--1643, St. Julian{'}s, Malta. Association for Computational Linguistics.

\bibitem[{Mohebbi et~al.(2023{\natexlab{a}})Mohebbi, Chrupa{\l}a, Zuidema, and Alishahi}]{mohebbi-etal-2023-homophone}
Hosein Mohebbi, Grzegorz Chrupa{\l}a, Willem Zuidema, and Afra Alishahi. 2023{\natexlab{a}}.
\newblock \href {https://doi.org/10.18653/v1/2023.emnlp-main.513} {Homophone disambiguation reveals patterns of context mixing in speech transformers}.
\newblock In \emph{Proceedings of the 2023 Conference on Empirical Methods in Natural Language Processing}, pages 8249--8260, Singapore. Association for Computational Linguistics.

\bibitem[{Mohebbi et~al.(2023{\natexlab{b}})Mohebbi, Zuidema, Chrupa{\l}a, and Alishahi}]{mohebbi-etal-2023-quantifying}
Hosein Mohebbi, Willem Zuidema, Grzegorz Chrupa{\l}a, and Afra Alishahi. 2023{\natexlab{b}}.
\newblock \href {https://doi.org/10.18653/v1/2023.eacl-main.245} {Quantifying context mixing in transformers}.
\newblock In \emph{Proceedings of the 17th Conference of the European Chapter of the Association for Computational Linguistics}, pages 3378--3400, Dubrovnik, Croatia. Association for Computational Linguistics.

\bibitem[{Moryossef et~al.(2019)Moryossef, Aharoni, and Goldberg}]{moryossef2019filling}
Amit Moryossef, Roee Aharoni, and Yoav Goldberg. 2019.
\newblock Filling gender \& number gaps in neural machine translation with black-box context injection.
\newblock In \emph{Proceedings of the First Workshop on Gender Bias in Natural Language Processing}, pages 49--54.

\bibitem[{Ramesh et~al.(2021)Ramesh, Gupta, and Singh}]{ramesh2021evaluating}
Krithika Ramesh, Gauri Gupta, and Sanjay Singh. 2021.
\newblock Evaluating gender bias in hindi-english machine translation.
\newblock In \emph{Proceedings of the 3rd Workshop on Gender Bias in Natural Language Processing}, pages 16--23.

\bibitem[{Reif et~al.(2019)Reif, Yuan, Wattenberg, Viegas, Coenen, Pearce, and Kim}]{reif2019}
Emily Reif, Ann Yuan, Martin Wattenberg, Fernanda~B. Viegas, Andy Coenen, Adam Pearce, and Been Kim. 2019.
\newblock Visualizing and measuring the geometry of bert.
\newblock In \emph{Advances in Neural Information Processing Systems (NeurIPS)}, pages 8594--8603.

\bibitem[{Rescigno et~al.(2020)Rescigno, Vanmassenhove, Monti, and Way}]{rescigno2020case}
Argentina~Anna Rescigno, Eva Vanmassenhove, Johanna Monti, and Andy Way. 2020.
\newblock A case study of natural gender phenomena in translation a comparison of google translate, bing microsoft translator and deepl for english to italian, french and spanish.
\newblock \emph{Computational Linguistics CLiC-it 2020}, page 359.

\bibitem[{Rios~Gonzales et~al.(2017)Rios~Gonzales, Mascarell, and Sennrich}]{rios-gonzales-etal-2017-improving}
Annette Rios~Gonzales, Laura Mascarell, and Rico Sennrich. 2017.
\newblock \href {https://doi.org/10.18653/v1/W17-4702} {Improving word sense disambiguation in neural machine translation with sense embeddings}.
\newblock In \emph{Proceedings of the Second Conference on Machine Translation}, pages 11--19, Copenhagen, Denmark. Association for Computational Linguistics.

\bibitem[{Rockt{\"a}schel et~al.(2016)Rockt{\"a}schel, Grefenstette, Hermann, Ko{\v{c}}isk{\'y}, and Blunsom}]{rocktaschel2016}
Tim Rockt{\"a}schel, Edward Grefenstette, Karl~Moritz Hermann, Tom{\'a}{\v{s}} Ko{\v{c}}isk{\'y}, and Phil Blunsom. 2016.
\newblock Reasoning about entailment with neural attention.
\newblock In \emph{International Conference on Learning Representations (ICLR)}.

\bibitem[{Rudinger et~al.(2018)Rudinger, Naradowsky, Leonard, and Van~Durme}]{rudinger-etal-2018-gender}
Rachel Rudinger, Jason Naradowsky, Brian Leonard, and Benjamin Van~Durme. 2018.
\newblock \href {https://doi.org/10.18653/v1/N18-2002} {Gender bias in coreference resolution}.
\newblock In \emph{Proceedings of the 2018 Conference of the North {A}merican Chapter of the Association for Computational Linguistics: Human Language Technologies, Volume 2 (Short Papers)}, pages 8--14, New Orleans, Louisiana. Association for Computational Linguistics.

\bibitem[{Sarti et~al.(2024)Sarti, Chrupała, Nissim, and Bisazza}]{sarti2024_context}
Gabriele Sarti, Grzegorz Chrupała, Malvina Nissim, and Arianna Bisazza. 2024.
\newblock \href {https://arxiv.org/abs/2310.01188} {Quantifying the plausibility of context reliance in neural machine translation}.
\newblock In \emph{Proceedings of the Twelfth International Conference on Learning Representations (ICLR)}.

\bibitem[{Saunders and Byrne(2020)}]{saunders2020reducing}
Danielle Saunders and Bill Byrne. 2020.
\newblock Reducing gender bias in neural machine translation as a domain adaptation problem.
\newblock In \emph{Proceedings of the 58th Annual Meeting of the Association for Computational Linguistics}, pages 7724--7736.

\bibitem[{Savoldi et~al.(2024)Savoldi, Bastings, Bentivogli, and Vanmassenhove}]{review_gender_bias_mt}
Beatrice Savoldi, Jasmijn Bastings, Lucia Bentivogli, and Eva Vanmassenhove. 2024.
\newblock A decade of gender bias in machine translation.
\newblock \emph{[under review]}.

\bibitem[{Savoldi et~al.(2021)Savoldi, Gaido, Bentivogli, Negri, and Turchi}]{savoldi2021gender}
Beatrice Savoldi, Marco Gaido, Luisa Bentivogli, Matteo Negri, and Marco Turchi. 2021.
\newblock Gender bias in machine translation.
\newblock \emph{Transactions of the Association for Computational Linguistics}, 9:845--874.

\bibitem[{Sennrich(2017)}]{sennrich-2017-grammatical}
Rico Sennrich. 2017.
\newblock \href {https://aclanthology.org/E17-2060/} {How grammatical is character-level neural machine translation? assessing {MT} quality with contrastive translation pairs}.
\newblock In \emph{Proceedings of the 15th Conference of the {E}uropean Chapter of the Association for Computational Linguistics: Volume 2, Short Papers}, pages 376--382, Valencia, Spain. Association for Computational Linguistics.

\bibitem[{Stahlberg et~al.(2007)Stahlberg, Braun, Irmen, and Sczesny}]{stahlberg}
Dagmar Stahlberg, Friederike Braun, Lisa Irmen, and Sabine Sczesny. 2007.
\newblock \href {http://psycnet.apa.org/psycinfo/2007-01308-006} {\emph{Representation of the sexes in language}}, page 163–187.

\bibitem[{Stanczak and Augenstein(2021)}]{stanczak2021survey}
Karolina Stanczak and Isabelle Augenstein. 2021.
\newblock {A survey on gender bias in natural language processing}.
\newblock \emph{arXiv preprint arXiv:2112.14168}.

\bibitem[{Stanovsky et~al.(2019)Stanovsky, Smith, and Zettlemoyer}]{winomt}
Gabriel Stanovsky, Noah~A. Smith, and Luke Zettlemoyer. 2019.
\newblock \href {https://doi.org/10.18653/v1/p19-1164} {Evaluating gender bias in machine translation}.
\newblock \emph{Proceedings of the 57th Annual Meeting of the Association for Computational Linguistics}.

\bibitem[{Sun et~al.(2019)Sun, Gaut, Tang, Huang, ElSherief, Zhao, Mirza, Belding, Chang, and Wang}]{sun-etal-2019-mitigating}
Tony Sun, Andrew Gaut, Shirlyn Tang, Yuxin Huang, Mai ElSherief, Jieyu Zhao, Diba Mirza, Elizabeth Belding, Kai-Wei Chang, and William~Yang Wang. 2019.
\newblock \href {https://doi.org/10.18653/v1/P19-1159} {Mitigating gender bias in natural language processing: Literature review}.
\newblock In \emph{Proceedings of the 57th Annual Meeting of the Association for Computational Linguistics}, pages 1630--1640, Florence, Italy. Association for Computational Linguistics.

\bibitem[{Sun et~al.(2021)Sun, Webster, Shah, Wang, and Johnson}]{sun2021they}
Tony Sun, Kellie Webster, Apu Shah, William~Yang Wang, and Melvin Johnson. 2021.
\newblock They, them, theirs: Rewriting with gender-neutral english.
\newblock \emph{arXiv preprint arXiv:2102.06788}.

\bibitem[{Tiedemann et~al.(2023)Tiedemann, Aulamo, Bakshandaeva, Boggia, Gr{\"o}nroos, Nieminen, Raganato, Scherrer, V{\'a}zquez, and Virpioja}]{opus_mt}
J{\"o}rg Tiedemann, Mikko Aulamo, Daria Bakshandaeva, Michele Boggia, {Stig Arne} Gr{\"o}nroos, Tommi Nieminen, Alessandro Raganato, Yves Scherrer, Ra{\'u}l V{\'a}zquez, and Sami Virpioja. 2023.
\newblock \href {https://doi.org/10.1007/s10579-023-09704-w} {Democratizing neural machine translation with opus-mt}.
\newblock \emph{Language Resources and Evaluation}.

\bibitem[{Tiedemann(2012)}]{opus_data}
Jörg Tiedemann. 2012.
\newblock Parallel data, tools and interfaces in {OPUS}.
\newblock In \emph{Proceedings of the Eight International Conference on Language Resources and Evaluation (LREC'12)}, Istanbul, Turkey. European Language Resources Association (ELRA).

\bibitem[{Union(2024)}]{EU_AI_ACT_2024}
European Union. 2024.
\newblock \href {https://eur-lex.europa.eu/eli/reg/2024/1689/oj} {Regulation (eu) 2024/1689 of the european parliament and of the council of 13 june 2024 laying down harmonised rules on artificial intelligence}.

\bibitem[{Vamvas and Sennrich(2021)}]{vamvas-sennrich-2021-contrastive}
Jannis Vamvas and Rico Sennrich. 2021.
\newblock \href {https://doi.org/10.18653/v1/2021.emnlp-main.803} {Contrastive conditioning for assessing disambiguation in {MT}: {A} case study of distilled bias}.
\newblock In \emph{Proceedings of the 2021 Conference on Empirical Methods in Natural Language Processing}, pages 10246--10265, Online and Punta Cana, Dominican Republic. Association for Computational Linguistics.

\bibitem[{Vamvas and Sennrich(2022)}]{vamvas-sennrich-2022-little}
Jannis Vamvas and Rico Sennrich. 2022.
\newblock \href {https://doi.org/10.18653/v1/2022.acl-short.53} {As little as possible, as much as necessary: Detecting over- and undertranslations with contrastive conditioning}.
\newblock In \emph{Proceedings of the 60th Annual Meeting of the Association for Computational Linguistics (Volume 2: Short Papers)}, pages 490--500, Dublin, Ireland. Association for Computational Linguistics.

\bibitem[{Van Der~Wal et~al.(2022)Van Der~Wal, Jumelet, Schulz, and Zuidema}]{birth_of_bias}
Oskar Van Der~Wal, Jaap Jumelet, Katrin Schulz, and Willem Zuidema. 2022.
\newblock \href {https://aclanthology.org/2022.gebnlp-1.8} {The birth of bias: A case study on the evolution of gender bias in an {E}nglish language model}.
\newblock In \emph{Proceedings of the 4th Workshop on Gender Bias in Natural Language Processing (GeBNLP)}, pages 75--75. Association for Computational Linguistics.

\bibitem[{Vanmassenhove(2024)}]{llm_bias_4}
Eva Vanmassenhove. 2024.
\newblock Gender bias in machine translation and the era of large language models.
\newblock \emph{Gendered Technology in Translation and Interpreting: Centering Rights in the Development of Language Technology}, page 225.

\bibitem[{Vanmassenhove et~al.(2021{\natexlab{a}})Vanmassenhove, Emmery, and Shterionov}]{vanmassenhove2021neutral}
Eva Vanmassenhove, Chris Emmery, and Dimitar Shterionov. 2021{\natexlab{a}}.
\newblock Neutral rewriter: A rule-based and neural approach to automatic rewriting into gender-neutral alternatives.
\newblock \emph{arXiv preprint arXiv:2109.06105}.

\bibitem[{Vanmassenhove et~al.(2018)Vanmassenhove, Hardmeier, and Way}]{vanmassenhove-etal-2018-getting}
Eva Vanmassenhove, Christian Hardmeier, and Andy Way. 2018.
\newblock \href {https://doi.org/10.18653/v1/D18-1334} {Getting gender right in neural machine translation}.
\newblock In \emph{Proceedings of the 2018 Conference on Empirical Methods in Natural Language Processing}, pages 3003--3008, Brussels, Belgium. Association for Computational Linguistics.

\bibitem[{Vanmassenhove et~al.(2021{\natexlab{b}})Vanmassenhove, Shterionov, and Gwilliam}]{mtese}
Eva Vanmassenhove, Dimitar Shterionov, and Matthew Gwilliam. 2021{\natexlab{b}}.
\newblock Machine translationese: Effects of algorithmic bias on linguistic complexity in machine translation.
\newblock In \emph{Proceedings of the 16th Conference of the European Chapter of the Association for Computational Linguistics: Main Volume}, pages 2203--2213.

\bibitem[{Vanmassenhove et~al.(2019)Vanmassenhove, Shterionov, and Way}]{vanmassenhove2019lost}
Eva Vanmassenhove, Dimitar Shterionov, and Andy Way. 2019.
\newblock Lost in translation: Loss and decay of linguistic richness in machine translation.
\newblock In \emph{Proceedings of Machine Translation Summit XVII: Research Track}, pages 222--232.

\bibitem[{Vaswani et~al.(2017)Vaswani, Shazeer, Parmar, Uszkoreit, Jones, Gomez, Kaiser, and Polosukhin}]{attention2017}
Ashish Vaswani, Noam Shazeer, Niki Parmar, Jakob Uszkoreit, Llion Jones, Aidan~N Gomez, \L~ukasz Kaiser, and Illia Polosukhin. 2017.
\newblock Attention is all you need.
\newblock In \emph{Advances in Neural Information Processing Systems}, volume~30. Curran Associates, Inc.

\bibitem[{Vig et~al.(2020)Vig, Gehrmann, Belinkov, Qian, Nevo, Sakenis, Huang, Singer, and Shieber}]{vig2020cma}
Jesse Vig, Sebastian Gehrmann, Yonatan Belinkov, Sharon Qian, Daniel Nevo, Simas Sakenis, Jason Huang, Yaron Singer, and Stuart Shieber. 2020.
\newblock \href {https://arxiv.org/abs/2004.12265} {Causal mediation analysis for interpreting neural nlp: The case of gender bias}.
\newblock \emph{Preprint}, arXiv:2004.12265.

\bibitem[{Voita et~al.(2021)Voita, Sennrich, and Titov}]{voita-etal-2021-analyzing}
Elena Voita, Rico Sennrich, and Ivan Titov. 2021.
\newblock \href {https://doi.org/10.18653/v1/2021.acl-long.91} {Analyzing the source and target contributions to predictions in neural machine translation}.
\newblock In \emph{Proceedings of the 59th Annual Meeting of the Association for Computational Linguistics and the 11th International Joint Conference on Natural Language Processing (Volume 1: Long Papers)}, pages 1126--1140, Online. Association for Computational Linguistics.

\bibitem[{Voita et~al.(2019)Voita, Talbot, Moiseev, Sennrich, and Titov}]{voita-etal-2019-analyzing}
Elena Voita, David Talbot, Fedor Moiseev, Rico Sennrich, and Ivan Titov. 2019.
\newblock \href {https://doi.org/10.18653/v1/P19-1580} {Analyzing multi-head self-attention: Specialized heads do the heavy lifting, the rest can be pruned}.
\newblock In \emph{Proceedings of the 57th Annual Meeting of the Association for Computational Linguistics}, pages 5797--5808, Florence, Italy. Association for Computational Linguistics.

\bibitem[{Wang et~al.(2016)Wang, Huang, Zhao et~al.}]{wang2016}
Yequan Wang, Minlie Huang, Li~Zhao, et~al. 2016.
\newblock Attention-based lstm for aspect-level sentiment classification.
\newblock In \emph{Proceedings of the 2016 Conference on Empirical Methods in Natural Language Processing (EMNLP)}, pages 606--615.

\bibitem[{Xu et~al.(2015)Xu, Ba, Kiros, Cho, Courville, Salakhudinov, Zemel, and Bengio}]{xu2015}
Kelvin Xu, Jimmy Ba, Ryan Kiros, Kyunghyun Cho, Aaron Courville, Ruslan Salakhudinov, Rich Zemel, and Yoshua Bengio. 2015.
\newblock Show, attend and tell: Neural image caption generation with visual attention.
\newblock In \emph{Proceedings of the International Conference on Machine Learning (ICML)}, pages 2048--2057.

\bibitem[{Yin et~al.(2021)Yin, Fernandes, Pruthi, Chaudhary, Martins, and Neubig}]{yin-etal-2021-context}
Kayo Yin, Patrick Fernandes, Danish Pruthi, Aditi Chaudhary, Andr{\'e} F.~T. Martins, and Graham Neubig. 2021.
\newblock \href {https://doi.org/10.18653/v1/2021.acl-long.65} {Do context-aware translation models pay the right attention?}
\newblock In \emph{Proceedings of the 59th Annual Meeting of the Association for Computational Linguistics and the 11th International Joint Conference on Natural Language Processing (Volume 1: Long Papers)}, pages 788--801, Online. Association for Computational Linguistics.

\bibitem[{Zhao et~al.(2018)Zhao, Wang, Yatskar, Ordonez, and Chang}]{zhao-etal-2018-gender}
Jieyu Zhao, Tianlu Wang, Mark Yatskar, Vicente Ordonez, and Kai-Wei Chang. 2018.
\newblock \href {https://doi.org/10.18653/v1/N18-2003} {Gender bias in coreference resolution: Evaluation and debiasing methods}.
\newblock In \emph{Proceedings of the 2018 Conference of the North {A}merican Chapter of the Association for Computational Linguistics: Human Language Technologies, Volume 2 (Short Papers)}, pages 15--20, New Orleans, Louisiana. Association for Computational Linguistics.

\bibitem[{Zhao et~al.(2024)Zhao, Ding, Jia, Wang, and Qian}]{gender_bias_llms2}
Jinman Zhao, Yitian Ding, Chen Jia, Yining Wang, and Zifan Qian. 2024.
\newblock \href {https://arxiv.org/abs/2403.00277} {Gender bias in large language models across multiple languages}.
\newblock \emph{arXiv}.

\bibitem[{Zmigrod et~al.(2019)Zmigrod, Mielke, Wallach, and Cotterell}]{zmigrod2019counterfactual}
Ran Zmigrod, Sabrina~J Mielke, Hanna Wallach, and Ryan Cotterell. 2019.
\newblock Counterfactual data augmentation for mitigating gender stereotypes in languages with rich morphology.
\newblock In \emph{Proceedings of the 57th Annual Meeting of the Association for Computational Linguistics}, pages 1651--1661.

\end{thebibliography}
